\documentclass{article} % For LaTeX2e
\usepackage[preprint]{colm2026_conference}

\usepackage{microtype}
\usepackage{hyperref}
\usepackage{url}
\usepackage{booktabs}
\usepackage{tabularx}
\usepackage{array}
\usepackage{xcolor}
\usepackage[dvipsnames]{xcolor}
\usepackage[table]{xcolor}
\usepackage{amsmath} 
\usepackage[most]{tcolorbox}
\usepackage{lineno}

\definecolor{darkblue}{rgb}{0, 0, 0.5}
\hypersetup{colorlinks=true, citecolor=darkblue, linkcolor=darkblue, urlcolor=darkblue}
\usepackage{graphicx}
\usepackage{multirow}
\usepackage{subcaption}  % 子图支持
\usepackage{tcolorbox}
\newtcolorbox{insightbox}{
    colback=blue!5!white, % 极浅的蓝色背景，非常学术
    colframe=blue!75!black, % 边框颜色
    left=4pt, right=4pt, top=4pt, bottom=4pt, % 极度压缩内部留白，省空间
    boxrule=1pt, % 边框粗细
    arc=3pt, % 圆角
    width=\columnwidth % 和正文同宽
}

\title{On the Role of Reasoning Patterns in the Generalization Discrepancy of Long Chain-of-Thought Supervised Fine-Tuning}
\author{Zhaoyi Li$^{1,2,3}$\thanks{Equal contributions}, Xiangyu Xi$^{2*}$\thanks{Corresponding authors: xixy10@foxmail.com, ying.wei@zju.edu.cn, liandefu@ustc.edu.cn.}, Zhengyu Chen$^{2}$, Wei Wang$^{2}$, 
\\
\textbf{Gangwei Jiang}$^{1,3}$, \textbf{Ranran Shen}$^{1}$, \textbf{Linqi Song}$^{3}$, \textbf{Ying Wei}$^{4\dagger}$, \textbf{Defu Lian}$^{1\dagger}$\\
$^{1}$University of Science and Technology of China, $^{2}$Meituan LongCat Team,
\\
$^{3}$City University of Hong Kong, $^{4}$Zhejiang University
}

\begin{document}

\ifcolmsubmission
\linenumbers
\fi

\maketitle

\begin{abstract}

Supervised Fine-Tuning (SFT) on long Chain-of-Thought (CoT) trajectories has become a pivotal phase in building large reasoning models.
However, how CoT trajectories from different sources influence the generalization performance of models remains an open question.
In this paper, we conduct a comparative study using two sources of verified CoT trajectories generated by two competing models, \texttt{DeepSeek-R1-0528} and \texttt{gpt-oss-120b}, with their problem sets controlled to be identical.
Despite their comparable performance, we uncover a striking paradox: lower training loss does not translate to better generalization.
SFT on \texttt{DeepSeek-R1-0528} data achieves remarkably lower training loss, yet exhibits significantly worse generalization performance on reasoning benchmarks compared to those trained on \texttt{gpt-oss-120b}. 
To understand this paradox, we perform a multi-faceted analysis probing token-level SFT loss and step-level reasoning behaviors.  
Our analysis reveals a difference in reasoning patterns. \texttt{gpt-oss-120b} exhibits highly convergent and deductive trajectories, whereas \texttt{DeepSeek-R1-0528} favors a divergent and branch-heavy exploration pattern.
Consequently, models trained with \texttt{DeepSeek-R1} data inherit inefficient exploration behaviors, often getting trapped in redundant exploratory branches that hinder them from reaching correct solutions.
Building upon this insight, we propose a simple yet effective remedy of filtering out frequently branching trajectories to improve the generalization of SFT. 
Experiments show that training on selected \texttt{DeepSeek-R1-0528} subsets surprisingly improves reasoning performance by up to $5.1\%$ on AIME25, $5.5\%$ on BeyondAIME, and on average $3.6\%$ on five benchmarks. 
\end{abstract}

\section{Introduction}
\label{sec:intro}
Supervised Fine-Tuning (SFT) on long Chain-of-Thought (CoT) trajectories has become a standard and foundational phase in building Large Reasoning Models (LRMs)~\citep{yang2025demystifying,lim2025motif}. 
%Under the standard paradigm, this process 
This common practice oftentimes initiates from 
%typically involves 
sampling extensive reasoning trajectories from existing LRMs (teacher models) and verifying their correctness. 
These trajectories are subsequently fed into base models (student models) in the cold-start SFT phase~\citep{guo2025deepseek,abdin2025phi4reasoningtechnicalreport} to endow them with complex problem-solving capabilities.
Building on this, a central question concerns \emph{which LRM should be used to generate such reasoning trajectories}. 
%A natural criterion is the selection of teacher models with strong reasoning performance, under the assumption that better-performing teachers provide more effective CoT data.
A natural idea is to investigate which LRM's trajectories the student models fit the best during SFT~\citep{tian2025not, zhang2025the}.

%To examine this assumption, we conduct a controlled comparative study using two LRMs with comparable performance. 
% We utilize \textit{verified, correct} reasoning trajectories elicited from two open-source, widely-adopted LRMs, \texttt{DeepSeek-R1-0528}~\citep{guo2025deepseek} and \texttt{gpt-oss-120b}~\citep{openai2025gptoss120bgptoss20bmodel} (in short, \texttt{R1} and \texttt{gpt-oss}), in response to an \textit{identical set} of $500$K complex mathematical problems. 
To examine this assumption, we conduct a controlled comparative SFT study utilizing \textit{\textbf{verified, correct}} reasoning trajectories elicited from two open-source, widely-adopted~\citep{zheng2025stabilizing,shmidman2025learning,yang2026reasoning} LRMs, \texttt{DeepSeek-R1-0528}~\citep{guo2025deepseek} and \texttt{gpt-oss-120b}~\citep{openai2025gptoss120bgptoss20bmodel} (in short, \texttt{R1} and \texttt{gpt-oss}), in response to an \textit{\textbf{identical set}} of $\sim500$K complex mathematical problems.
Through SFT across four base models (including different model families and scales), we observe a stark and consistent \textbf{generalization discrepancy}: training on \texttt{R1} data yields much \textit{\textbf{lower SFT training loss}}, yet it results in notably \textit{\textbf{inferior generalization performance}} on downstream reasoning benchmarks compared to \texttt{gpt-oss}. 
This naturally raises a research question: 
what properties of teacher-generated CoT trajectories truly affect their effectiveness for generalization ability?

% makes \texttt{R1} a less effective teacher model for Long CoT SFT compared to \texttt{gpt-oss}?

While recent studies~\citep{yang2026reasoning,panigrahi2026in} have highlighted the importance of  SFT data diversity and teacher-student distribution alignment in this process, %we find that 
these %surface-level 
metrics fail to explain the above generalization discrepancy with different teachers, %in our experiments, 
which motivates us to seek more %fundamental 
fine-grained differences.
% factors.
% Although 
Inspired by emerging works~\citep{minegishi2025topology,bogdan2025thought,jiang2025makes} %have begun to
that characterize the effectiveness of long CoT trajectories through %their 
inherent structural patterns, we delve into the specific role of these \textit{reasoning patterns} %embedded within different teacher models 
in explaining the generalization discrepancy of different teacher models. %student models after SFT remains an open question.

% To investigate this, we conduct a controlled comparative study. We utilize \textit{verified, correct} reasoning trajectories elicited from two open-source, widely-adopted LRMs, \texttt{DeepSeek-R1-0528}~\citep{guo2025deepseek} and \texttt{gpt-oss-120b}~\citep{openai2025gptoss120bgptoss20bmodel} (in short, \texttt{R1} and \texttt{gpt-oss}), in response to an \textit{identical set} of $500$K complex mathematical problems. 
% Through SFT across four base models (including different model families and scales), we observe a stark and consistent \textbf{generalization discrepancy}: training on \texttt{R1} data yields much \textit{\textbf{lower SFT training loss}}, yet it results in notably \textit{\textbf{inferior generalization performance}} on downstream reasoning benchmarks compared to \texttt{gpt-oss}. 
% This naturally raises a research question: since both training sets provide valid reasoning paths corresponding to the same set of problems, what makes \texttt{R1} a less effective teacher model for Long CoT SFT compared to \texttt{gpt-oss}?

% To understand this observation, 
Concretely, we perform a multi-faceted analysis encompassing both token-level SFT loss analysis and step-level reasoning behavior analysis~\citep{gandhi2025cognitive} for the training data generated by \texttt{R1} and \texttt{gpt-oss}. 
\emph{\textbf{At the token level}}, we observe 
%First, our token-level loss decomposition reveals 
a severe long-tail distribution, where the primary SFT optimization signals stem from a small subset of high-loss \textit{key reasoning tokens} indicating logical transitions. 
Although the average losses of these key tokens are quite close between \texttt{R1} and \texttt{gpt-oss}, \texttt{R1} data contains a much larger proportion of near-zero-loss tokens, explaining its lower overall SFT loss without necessarily improving on key reasoning signals.
\emph{\textbf{At the trajectory level}}, we uncover a fundamental difference in reasoning patterns 
%Furthermore, the semantic nature of these key tokens differs: \texttt{gpt-oss} favors convergent exploitation, whereas \texttt{R1} favors divergent exploration.
% Second, 
by automatically annotating each reasoning step into discrete reasoning behaviors using \texttt{DeepSeek-V3.2}~\citep{deepseekai2025deepseekv32pushingfrontieropen}. % we compare their distributions and transition matrices between \texttt{R1} and \texttt{gpt-oss}. 
%we uncover that 
\texttt{gpt-oss} focuses on deep and convergent deductive chains, whereas \texttt{R1} favors a divergent reasoning pattern characterized by \textit{\textbf{frequently branching into alternative ideas without deep deduction}}, which introduces numerous exploratory paths that are redundant to the main reasoning path. 

Crucially, we also demonstrate that reasoning patterns transfer to student models  through SFT. Student models trained with R1 data often get trapped in redundant exploratory branches that hinder them from reaching the correct solutions.
% Finally, 
We further validate such reasoning pattern and the transfer behavior through two intervention experiments. On one hand, 
%design an experiment in which 
we randomly remove part of the reasoning steps from each trajectory and retrain the models. 
While models trained on pruned \texttt{gpt-oss} data suffer severe performance drops, those trained on pruned \texttt{R1} data exhibit minimal degradation and, on some benchmarks, even gains.
On the other hand, we propose to improve the generalization of SFT with a targeted data curation strategy: \textbf{\textit{filtering out the most frequently branching trajectories from the \texttt{R1} dataset}} with two carefully designed proxy metrics.
We empirically demonstrate that \textit{\textbf{training on this reduced \texttt{R1} subset surprisingly improves reasoning performance}} by up to $5.1\%$ on AIME25, $5.5\%$ on BeyondAIME, and an average of $3.6\%$ on five benchmarks. 

In summary, our main contributions are three-fold:
(1) We expose a stark generalization discrepancy in long CoT SFT, revealing that trajectories from \texttt{R1} consistently induce lower training loss but notably worse generalization performance compared to \texttt{gpt-oss} under controlled settings.
(2) We demonstrate that this discrepancy is intrinsically correlated with differing reasoning patterns in training data: \texttt{R1} frequently branches into alternative reasoning paths without deep deduction, whereas \texttt{gpt-oss} provides more convergent deductive reasoning chains, which are inherited by student models through SFT.
and (3) We propose a simple yet effective approach to improve the generalization of SFT by filtering out the most frequently branching trajectories from the original \texttt{R1} dataset and empirically demonstrate that training on this reduced subset consistently improves reasoning performance.
\section{Related Work}
\label{sec:related_work}
\textbf{Distilling Reasoning Ability via Long CoT Supervised Fine-Tuning}
The advent of LRMs, such as OpenAI's o1 \citep{jaech2024openai} and DeepSeek-R1 \citep{guo2025deepseek}, has shifted the paradigm of complex problem-solving toward prolonged thinking with branching, reflection, and backtracking (i.e., Long CoT reasoning \citep{yang2025demystifying,kopiczko2026data}). 
SFT with reasoning trajectories output by LRMs has become a focal point of research for two critical purposes: distilling the complex reasoning capacities of stronger LRMs into weaker models \citep{shridhar2023distilling}, and establishing a robust cold-start foundation to bootstrap further RL training in new frontier LRMs \citep{zhang2025interplay, matsutani2026rl, li2026getting}. 
Recent literature \citep{lim2025motif,zheng2025stabilizing} has recognized that the effectiveness of long CoT SFT mainly depends on how to select appropriate teacher models. 
\citet{tian2025not} demonstrated that not all correct reasoning trajectories contribute equally to student performance. \citet{shmidman2025learning} showed empirical comparisons between different teacher traces have revealed discrepancies in downstream performance.
\citet{chandra2025shape}, \citet{jung2025prismatic}, \citet{panigrahi2026in}, and \citet{yang2026reasoning} emphasized the “teacher-student compatibility”: successful distillation requires striking a delicate balance among the diversity, informativeness, and distribution alignment of traces to optimally suit the student model.
In contrast, this work focuses on the intrinsic properties (reasoning patterns)~\citep{wang2025beyond,jiang2025makes,gandhi2025cognitive} of the reasoning trajectories and analyzes the role of these reasoning patterns in the generalization of SFT.

\textbf{Analyzing the Intrinsic Properties of Long CoT Reasoning Trajectories}
As the study of Long CoT progresses, recent literature has increasingly recognized that reasoning trajectories generated by different LRMs possess fundamentally distinct intrinsic properties and characteristic reasoning behaviors \citep{sun2025idiosyncrasies, chen2025your,ballon2026probing}, and different types of reasoning trajectories may exhibit varying degrees of learnability~\citep{prasad2026effective}.
Recent efforts \citep{xiong2025mapping,minegishi2025topology,li2025understanding,jiang2025makes,li2026cotjudgergraphdrivenframeworkautomatic} have conceptualized reasoning trajectories as graphs, characterizing the reasoning properties with graph properties like radius, number of loops, and other GNN-captured features~\citep{brody2022how}. 
Besides, micro-level token-related features have also drawn attention~\citep{qian2025demystifying,wang2025beyond,li2025attention,chen2026think}, showing that a small fraction of critical tokens often drives the actual cognitive progress. Furthermore, mapping these trajectories into discrete cognitive behaviors and analyzing their relative proportions and transition probabilities have proven vital for characterizing reasoning patterns \citep{gandhi2025cognitive, bogdan2025thought, chen2026molecular}.
Building upon these analytical foundations, our systematically investigate how the reasoning patterns of long CoT trajectories dictate the efficacy and generalization of Long CoT SFT.
\section{Comparing SFT with Trajectories from Two Advanced LRMs}
\label{sec:comparative_study}
\begin{table}[t]
% \vspace{-4mm}
\centering
\caption{SFT performance comparison of different base models trained on reasoning trajectories generated by \texttt{DeepSeek-R1} and \texttt{gpt-oss-120b} across five mathematical reasoning benchmarks. AIME24/25 results are averaged over 32 independent runs (\texttt{avg@32}). BeyondAIME and HMMT25 results are averaged over 10 independent runs (\texttt{avg@10}).}
\vspace{-2mm}
\label{tab:main_results_compare_r1_and_oss}
\resizebox{\textwidth}{!}{%
\begin{tabular}{@{}l|l|ccccc|c@{}}
\toprule
\textbf{Base Model} & \textbf{Data Source} & \textbf{MATH500} & \textbf{AIME24} & \textbf{AIME25} & \textbf{BeyondAIME} & \textbf{HMMT25} & \textbf{Avg} \\
\midrule
\multirow{2}{*}{Qwen2.5-7B} 
 & DeepSeek-R1   & $95.0\%$\scriptsize$\pm0.02\%$ & $52.1\%$\scriptsize$\pm0.03\%$ & $44.8\%$\scriptsize$\pm0.03\%$ & $16.7\%$\scriptsize$\pm0.02\%$ & $38.7\%$\scriptsize$\pm0.03\%$ & $49.5\%$ \\
 & gpt-oss-120b  & $95.0\%$\scriptsize$\pm0.02\%$ & $\mathbf{64.5\%}$\scriptsize$\pm0.03\%$ & $\mathbf{46.7\%}$\scriptsize$\pm0.03\%$ & $\mathbf{23.9\%}$\scriptsize$\pm0.03\%$ & $\mathbf{42.9\%}$\scriptsize$\pm0.03\%$ & $\mathbf{54.6\%}$ {\footnotesize \textcolor{green!60!black}{(\textbf{+5.1\%})}} \\
\midrule
\multirow{2}{*}{Qwen2.5-32B} 
 & DeepSeek-R1   & $98.4\%$\scriptsize$\pm0.01\%$ & $78.5\%$\scriptsize$\pm0.03\%$ & $73.5\%$\scriptsize$\pm0.03\%$ & $37.7\%$\scriptsize$\pm0.03\%$ & $69.6\%$\scriptsize$\pm0.03\%$ & $71.5\%$ \\
 & gpt-oss-120b  & $\mathbf{98.8\%}$\scriptsize$\pm0.01\%$ & $\mathbf{82.0\%}$\scriptsize$\pm0.02\%$ & $\mathbf{77.7\%}$\scriptsize$\pm0.03\%$ & $\mathbf{45.9\%}$\scriptsize$\pm0.03\%$ & $\mathbf{77.3\%}$\scriptsize$\pm0.03\%$ & $\mathbf{76.3\%}$ {\footnotesize \textcolor{green!60!black}{(\textbf{+4.8\%})}} \\
\midrule
\multirow{2}{*}{Llama3.1-8B} 
 & DeepSeek-R1   & $86.2\%$\scriptsize$\pm0.03\%$ & $20.1\%$\scriptsize$\pm0.03\%$ & $18.3\%$\scriptsize$\pm0.02\%$ & $4.5\%$\scriptsize$\pm0.01\%$  & $18.5\%$\scriptsize$\pm0.02\%$ & $29.5\%$ \\
 & gpt-oss-120b  & $\mathbf{94.8\%}$\scriptsize$\pm0.02\%$ & $\mathbf{53.8\%}$\scriptsize$\pm0.03\%$ & $\mathbf{41.3\%}$\scriptsize$\pm0.03\%$ & $\mathbf{21.8\%}$\scriptsize$\pm0.03\%$ & $\mathbf{40.6\%}$\scriptsize$\pm0.03\%$ & $\mathbf{50.5\%}$ {\footnotesize \textcolor{green!60!black}{(\textbf{+21.0\%})}} \\
\midrule
\multirow{2}{*}{Qwen3-8B} 
 & DeepSeek-R1   & $\mathbf{96.8\%}$\scriptsize$\pm0.02\%$ & $63.0\%$\scriptsize$\pm0.03\%$ & $54.0\%$\scriptsize$\pm0.03\%$ & $24.5\%$\scriptsize$\pm0.03\%$ & $48.5\%$\scriptsize$\pm0.03\%$ & $57.4\%$ \\
 & gpt-oss-120b  & $96.2\%$\scriptsize$\pm0.02\%$ & $\mathbf{70.6\%}$\scriptsize$\pm0.03\%$ & $\mathbf{58.9\%}$\scriptsize$\pm0.03\%$ & $\mathbf{32.5\%}$\scriptsize$\pm0.03\%$ & $\mathbf{55.3\%}$\scriptsize$\pm0.03\%$ & $\mathbf{62.7\%}$ {\footnotesize \textcolor{green!60!black}{(\textbf{+5.3\%})}} \\
\bottomrule
\end{tabular}%
}
\vspace{-4mm}
\end{table}
To uncover the underlying principles of long CoT SFT, we conduct a controlled comparative study, in which we utilize reasoning trajectories from two open-weight, widely-adopted, and advanced LRMs: \texttt{DeepSeek-R1}~\citep{guo2025deepseek} ("\texttt{DeepSeek-R1}" hereinafter refers to "\texttt{DeepSeek-R1-0528}"\footnote{\url{https://huggingface.co/deepseek-ai/DeepSeek-R1-0528}}) and \texttt{gpt-oss-120b}~\citep{openai2025gptoss120bgptoss20bmodel} to fine-tune base models. 
By ensuring that these trajectories \textbf{correspond to an identical set of prompts (questions)} and are \textbf{all verified to yield correct answers}~\citep{tian2025not,shmidman2025learning}, we eliminate confounding factors related to problem distribution or factual correctness, allowing us to empirically investigate how the distinct intrinsic properties of different teacher trajectories affect SFT and the resulting reasoning generalization of student base models.
Specifically, our experimental settings are as below:
\textbf{(1) SFT Data Collection:} Following~\citet{tian2025not}, we collect a high-quality dataset comprising approximately 500,000 challenging mathematical problems from publicly available datasets including OpenR1-Math-220k~\citep{openr1}, Big-Math-RL-Verified~\citep{albalak2025bigmathlargescalehighqualitymath}, NuminaMath~\citep{numina_math_datasets} and so on. For each problem, we query both \texttt{DeepSeek-R1} and \texttt{gpt-oss-120b} to generate their respective Long CoT trajectories. To control data quality, we apply a rule-based verification pipeline to ensure that \textit{all trajectories used in our experiments successfully arrive at the correct final answer}. 
The context length of the reasoning trajectories used for SFT is controlled below 32k;
\textbf{(2) Base Models:} We select four representative (covering different model families and scales) open-weight base models as student models: Qwen2.5-7B~\citep{qwen2025qwen25technicalreport}, Qwen2.5-32B~\citep{qwen2025qwen25technicalreport}, Llama3.1-8B~\citep{grattafiori2024llama3herdmodels}, and Qwen3-8B~\citep{yang2025qwen3technicalreport}. Each model undergoes SFT on the \texttt{DeepSeek-R1} and \texttt{gpt-oss-120b} datasets independently. We maintain identical hyperparameter settings and optimize for \textit{approximately the same number of training steps} across each pair of comparisons;
and \textbf{(3) Evaluation Benchmarks:} The resulting models are evaluated on five representative mathematical reasoning benchmarks: MATH500~\citep{hendrycks2021measuring}, AIME24~\citep{aime24}, AIME25~\citep{aime25}, BeyondAIME~\citep{bytedance_seed_2025_beyondaime}, and HMMT25~\citep{balunovic_srimatharena_2025}.
The inference context limit is set to 32k by default, aligned with the training data.
More details on our experimental setup can be found in Appendix~\ref{appendix:experiment_setup}.

\textbf{Empirical Observations on the SFT results.}
\label{subsec:observations}
\begin{figure}[t]
% \vspace{-6mm}
    \centering
    \captionsetup[subfigure]{justification=centering} 
    % 第一行
    \begin{subfigure}[t]{0.32\textwidth}
        \centering
        \includegraphics[width=\textwidth]{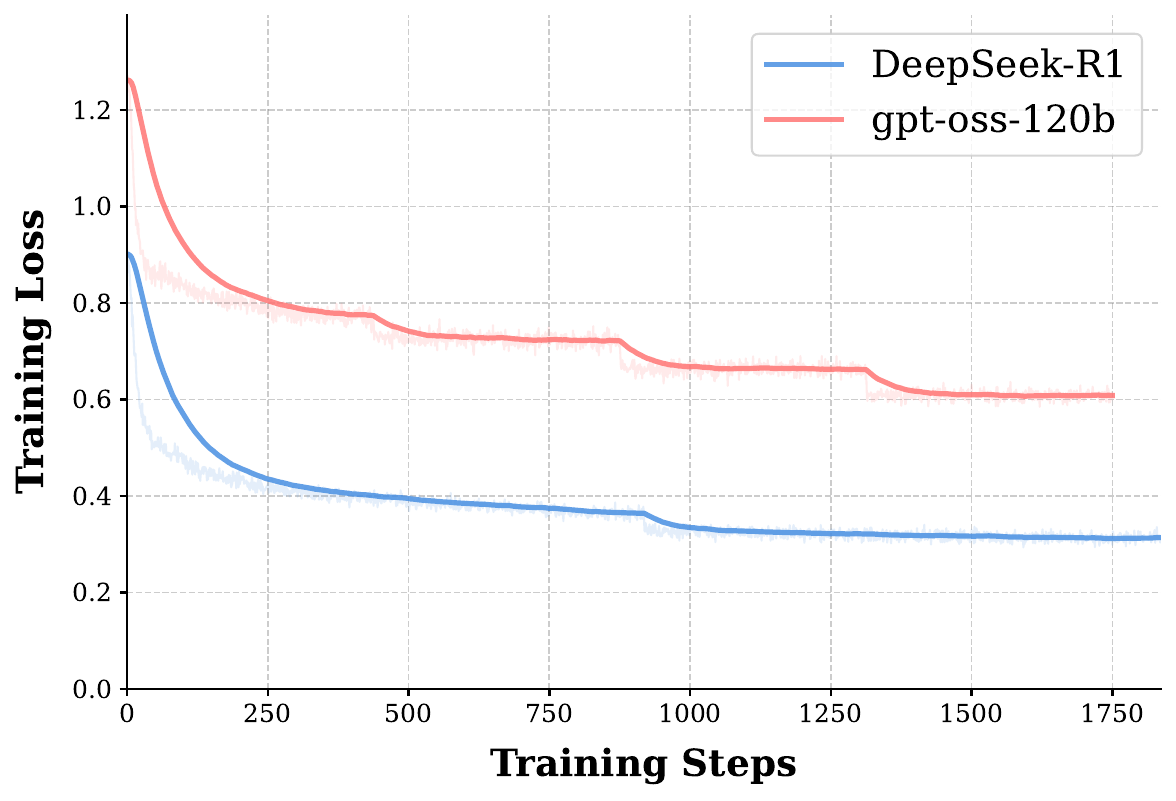}
        \caption{SFT training loss with\\Qwen2.5-7B}
    \end{subfigure}
    \begin{subfigure}[t]{0.32\textwidth}
        \centering
        \includegraphics[width=\textwidth]{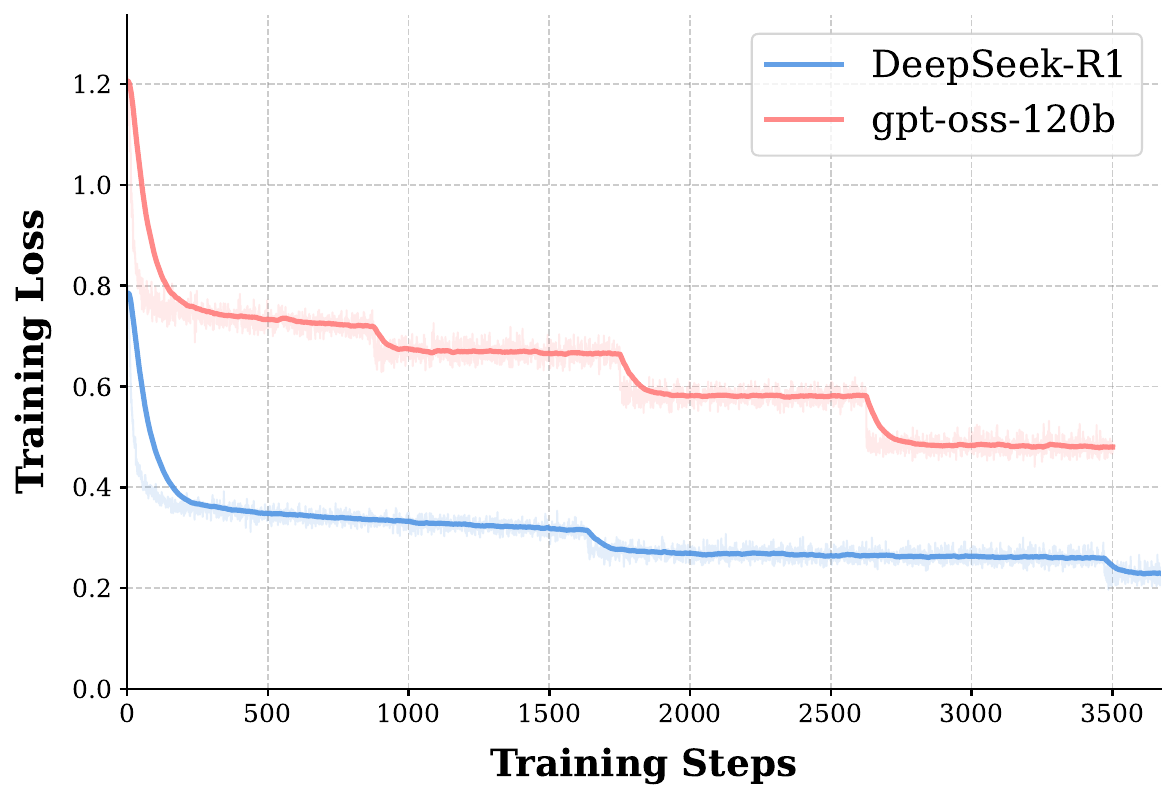}
        \caption{SFT training loss with\\Qwen2.5-32B}
    \end{subfigure}
    \begin{subfigure}[t]{0.32\textwidth}
        \centering
        \includegraphics[width=\textwidth]{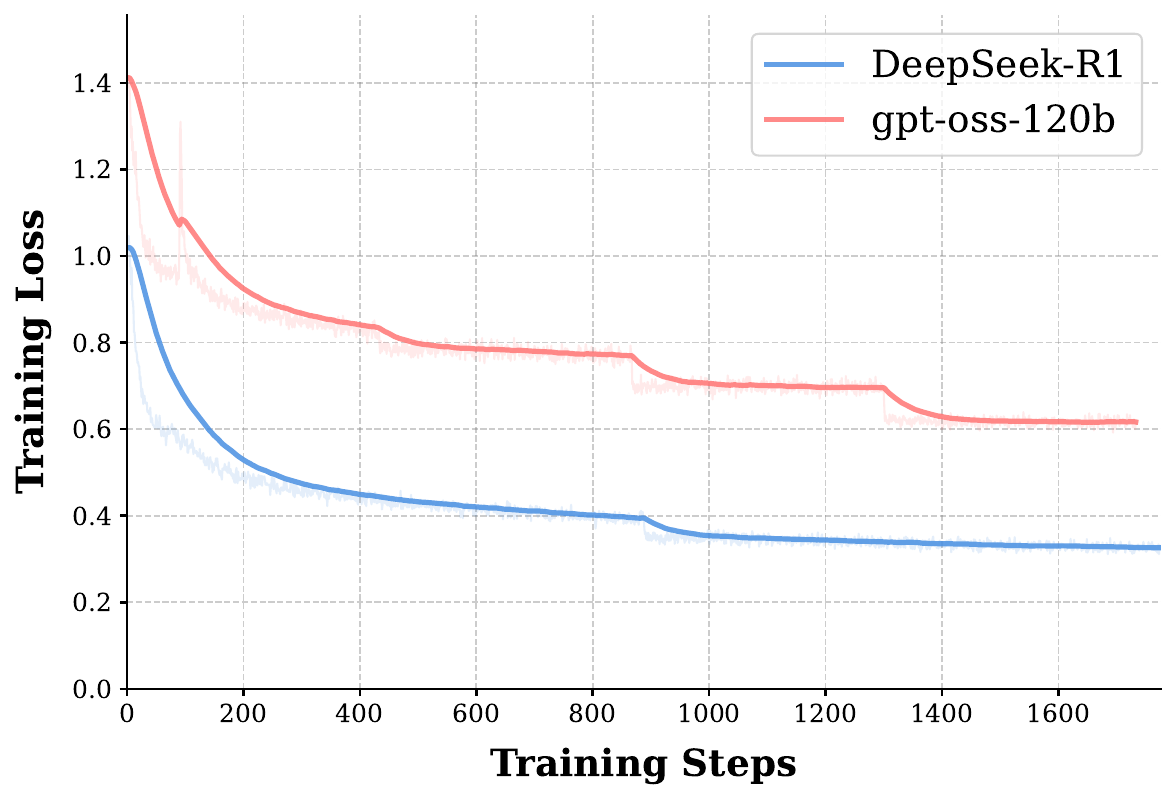}
        \caption{SFT training loss with\\Llama3.1-8B}
    \end{subfigure}
    \vspace{-2mm}
    \begin{subfigure}[t]{0.32\textwidth}
        \centering
        \includegraphics[width=\textwidth]{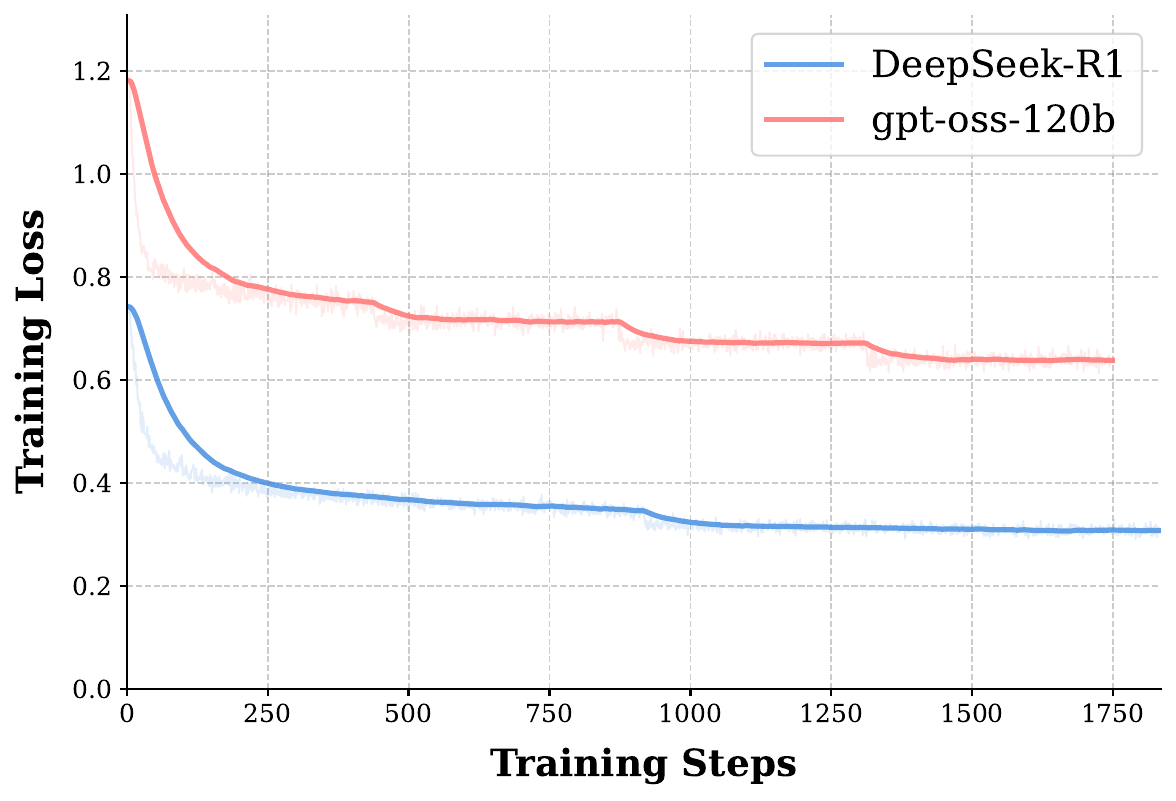}
        \caption{SFT training loss with\\Qwen3-8B}
    \end{subfigure}
    \begin{subfigure}[t]{0.32\textwidth}
        \centering
        \includegraphics[width=\textwidth]{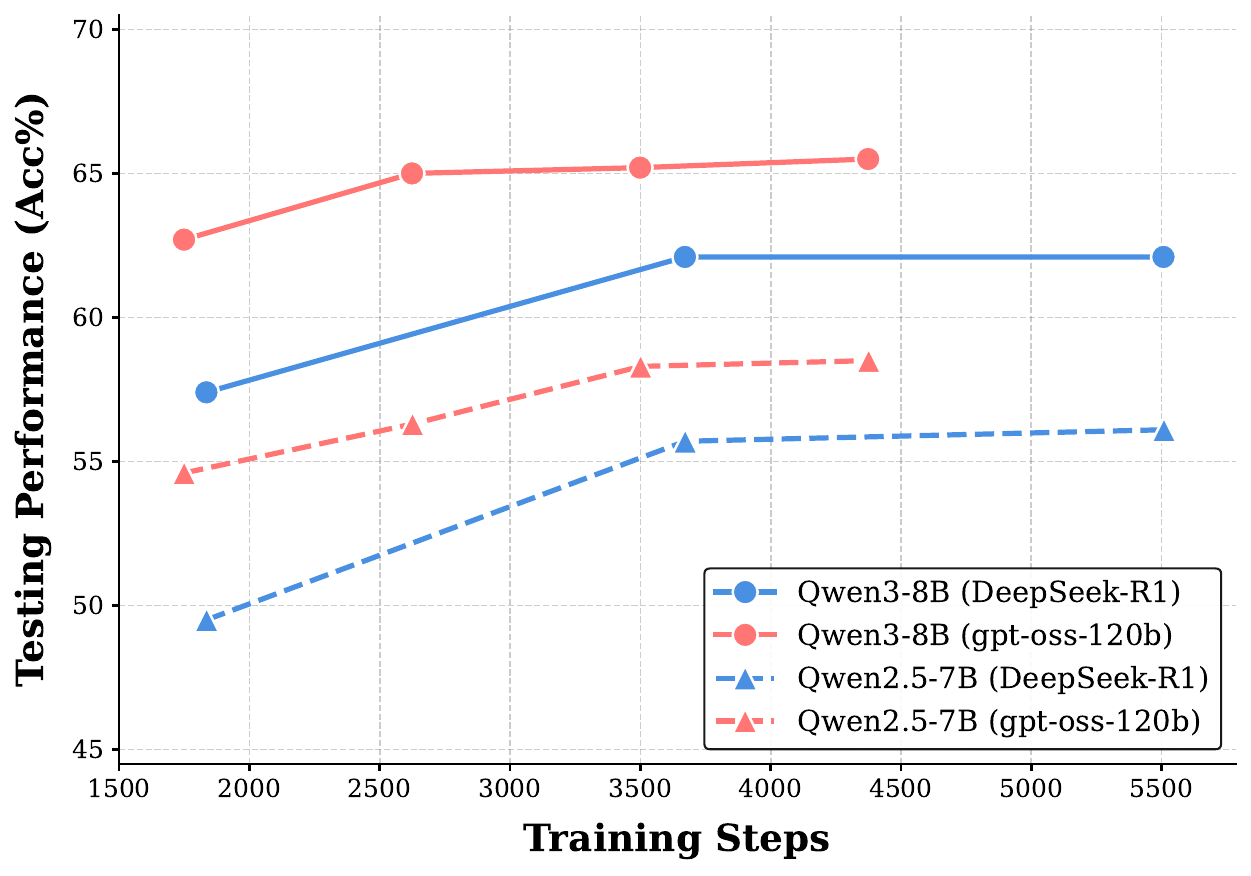}
        \caption{Testing performance with \\
        varying training steps}
    \end{subfigure}
    \begin{subfigure}[t]{0.32\textwidth}
        \centering
        \includegraphics[width=\textwidth]{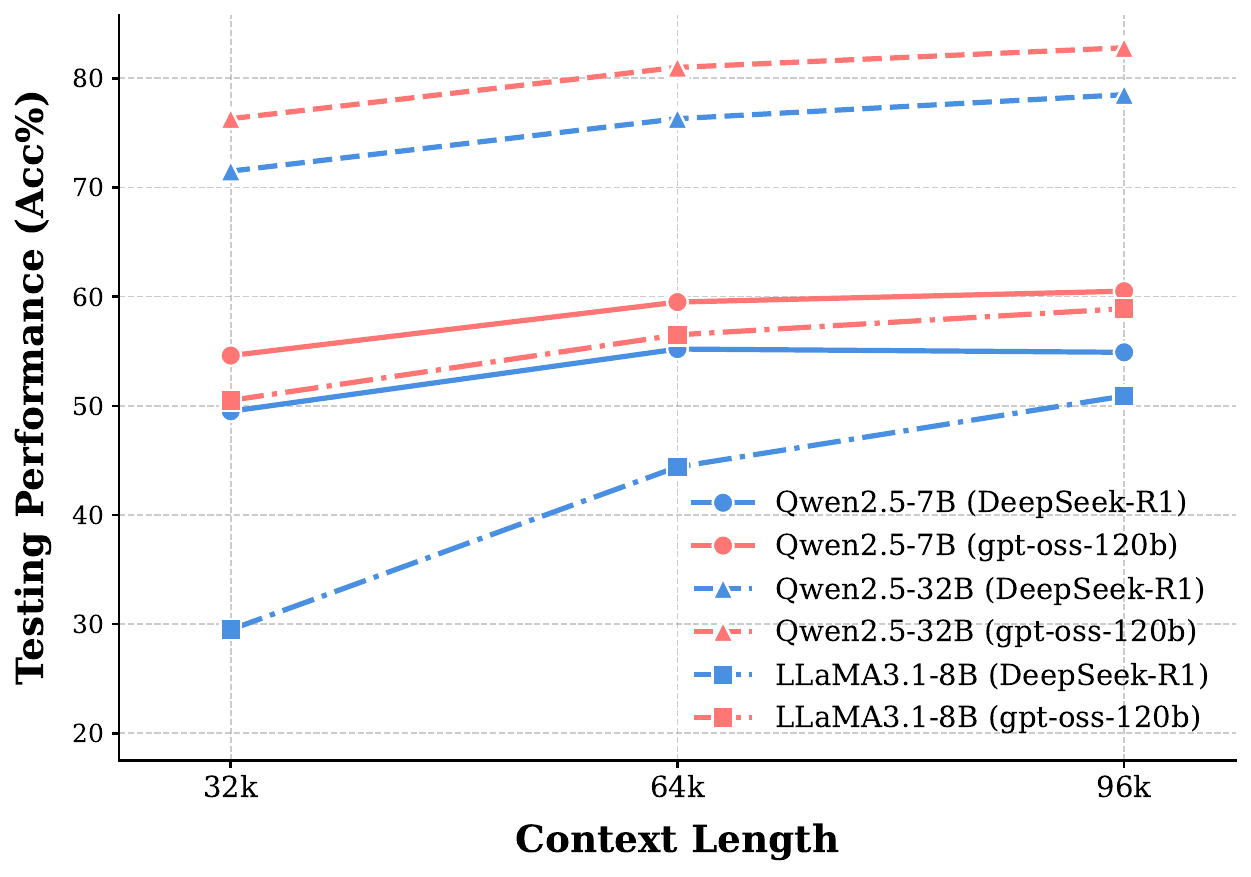}
        \caption{Testing performance with
        varying inference context length}
    \end{subfigure}
    \caption{(a $\sim$ d): SFT training loss comparison of different models trained on long CoT trajectories of \texttt{DeepSeek-R1} and \texttt{gpt-oss-120b}. (e) and (f): average testing performance on five benchmarks with varying training steps and inference context length. \textcolor{blue}{Blue}/\textcolor{red}{red} curves refer to experiments with \texttt{gpt-oss-120b}/\texttt{DeepSeek-R1}-generated data, respectively.}
    \label{fig:main_loss_compare_r1_and_oss}
    \vspace{-6mm}
\end{figure}
Based on the SFT results presented in Table \ref{tab:main_results_compare_r1_and_oss} and Figure \ref{fig:main_loss_compare_r1_and_oss}, we highlight two key observations:
\textbf{(1) The Generalization Discrepancy.} 
As detailed in Table \ref{tab:main_results_compare_r1_and_oss}, student models fine-tuned on \texttt{gpt-oss-120b} reasoning trajectories consistently and significantly outperform those trained on \texttt{DeepSeek-R1} data. 
This performance advantage holds across all four evaluated base models, regardless of model scale or architecture. 
This universal superiority is non-trivial: given that both training datasets cover the exact same set of prompts and strictly consist of trajectories that arrive at the correct final answers, the generalization performance discrepancy cannot be attributed to the factual correctness or the distribution of training data. 
This naturally raises a research question: \textit{\textbf{since both sets provide valid reasoning paths, what intrinsic factors make the consistent generalization discrepancy between models trained on \texttt{DeepSeek-R1} and \texttt{gpt-oss-120b} trajectories?}}
\textbf{(2) The SFT Loss Discrepancy.}
As an auxiliary finding, we observe an abnormal discrepancy in the training dynamics during SFT. 
As illustrated in Figure \ref{fig:main_loss_compare_r1_and_oss}(a $\sim$ d), across all student models, the SFT training loss on \texttt{DeepSeek-R1} data converges to a remarkably lower level (approximately $0.2 \sim 0.3$). In contrast, the training loss on \texttt{gpt-oss-120b} data plateaus at a significantly higher level (around $0.6$). 
In the SFT phase, a lower training loss typically correlates with better alignment to the target distribution, indicating a better learnability of reasoning trajectories~\citep{tian2025not,zhang2025the}. 
However, our experiments shows the \texttt{gpt-oss-120b} trajectories produce consistently superior student models although much ``harder'' to fit during training. 
This discrepancy is particularly pronounced in base models such as Llama3.1-8B, where the average accuracy drops from $50.5\%$ (\texttt{gpt-oss-120b}) to $29.5\%$ (\texttt{DeepSeek-R1}) despite the latter achieving a much lower training loss.

\textbf{Ruling out Confounding Factors.}
Before proceeding to the analysis of our observation, we conduct auxiliary experiments to ensure that our observations are not artifacts of the training or inference setups:
\textbf{(1) Extending Training Steps:} One might argue that the \texttt{DeepSeek-R1} data simply requires more training steps to converge. 
To test this, we vary the SFT training steps (from 1500 steps to 4500 steps) until the test-set performance fully saturates. 
The results for two base models (Qwen2.5-7B and Qwen3-8B) are shown in Figure~\ref{fig:main_loss_compare_r1_and_oss}(e), we find that the performance gap persists: the \texttt{gpt-oss-120b}-trained models still significantly and consistently outperform the \texttt{DeepSeek-R1}-trained models, confirming that the discrepancy is rooted in the intrinsic properties rather than undertraining;
and \textbf{(2) Scaling the Context Limit during Inference:} Long CoT reasoning trajectories from \texttt{DeepSeek-R1} are statistically longer than those from \texttt{gpt-oss-120b}. It is possible that \texttt{DeepSeek-R1}-trained student models perform poorly simply because their generations are prematurely truncated during evaluation~\citep{shmidman2025learning}. However, when we increase the maximum generation length (from 32k to 96k) for Qwen2.5-7B, Qwen2.5-32B and LLaMA3.1-8B (Qwen3-8B supports a maximum context length of 32k), at inference time, the \texttt{DeepSeek-R1}-trained students still fall behind \texttt{gpt-oss-120b}-trained ones, as depicted in Figure~\ref{fig:main_loss_compare_r1_and_oss} (f).
\begin{insightbox}
\textbf{Takeaway:}
under controlled settings, models trained on \texttt{DeepSeek-R1} data achieve much lower SFT loss, yet exhibit notably inferior generalization compared to \texttt{gpt-oss-120b}.
\end{insightbox}
\section{Comparing Reasoning Patterns in DeepSeek-R1 and gpt-oss-120b}
\label{sec:mechanistic_analysis}
To investigate the underlying causes of the empirical discrepancy in generalization and SFT loss observed in Section \ref{sec:comparative_study}, we perform a multi-faceted analysis. We first deconstruct the SFT loss at the token-level and subsequently formalize the structural-level pattern differences of reasoning trajectories through reasoning behavior analysis. 
We show that \texttt{DeepSeek-R1}'s reasoning trajectories exhibit a lower density of key reasoning tokens and favor a divergent exploration pattern with repetitive branching, resulting in a much higher degree of structural redundancy than \texttt{gpt-oss-120b}.
We then randomly delete reasoning steps from the original training data (from \texttt{DeepSeek-R1} and \texttt{gpt-oss-120b}) and re-train the base models with intervened data to further support our analysis. Due to the page limit, we only include part of the results here. More experiment results can be found in Appendix~\ref{appendix:more_experiment_results}.

\begin{figure}[b]
\vspace{-6mm}
    \centering
    \captionsetup[subfigure]{justification=centering} 
    % 第一行
    % \begin{subfigure}[t]{0.32\textwidth}
    %     \centering
    %     \includegraphics[width=\textwidth]{figure/loss_distribution_of_two_original_models.qwen3_8b.pdf}
    %     \caption{Qwen3-8B’s token-level loss distribution before SFT.}
    % \end{subfigure}
    % \begin{subfigure}[t]{0.32\textwidth}
    %     \centering
    %     \includegraphics[width=\textwidth]{figure/qwen3_8b_dpsk_r1_base_model_word_cloud.pdf}
    %     \caption{Most frequent tokens in the \texttt{DeepSeek-R1} data before SFT.}
    % \end{subfigure}
    % \begin{subfigure}[t]{0.32\textwidth}
    %     \centering
    %     \includegraphics[width=\textwidth]{figure/qwen3_8b_oss_120b_base_model_word_cloud.pdf}
    %     \caption{Most frequent tokens in the \texttt{gpt-oss-120b} data before SFT.}
    % \end{subfigure}
    \begin{subfigure}[t]{0.32\textwidth}
        \centering
        \includegraphics[width=\textwidth]{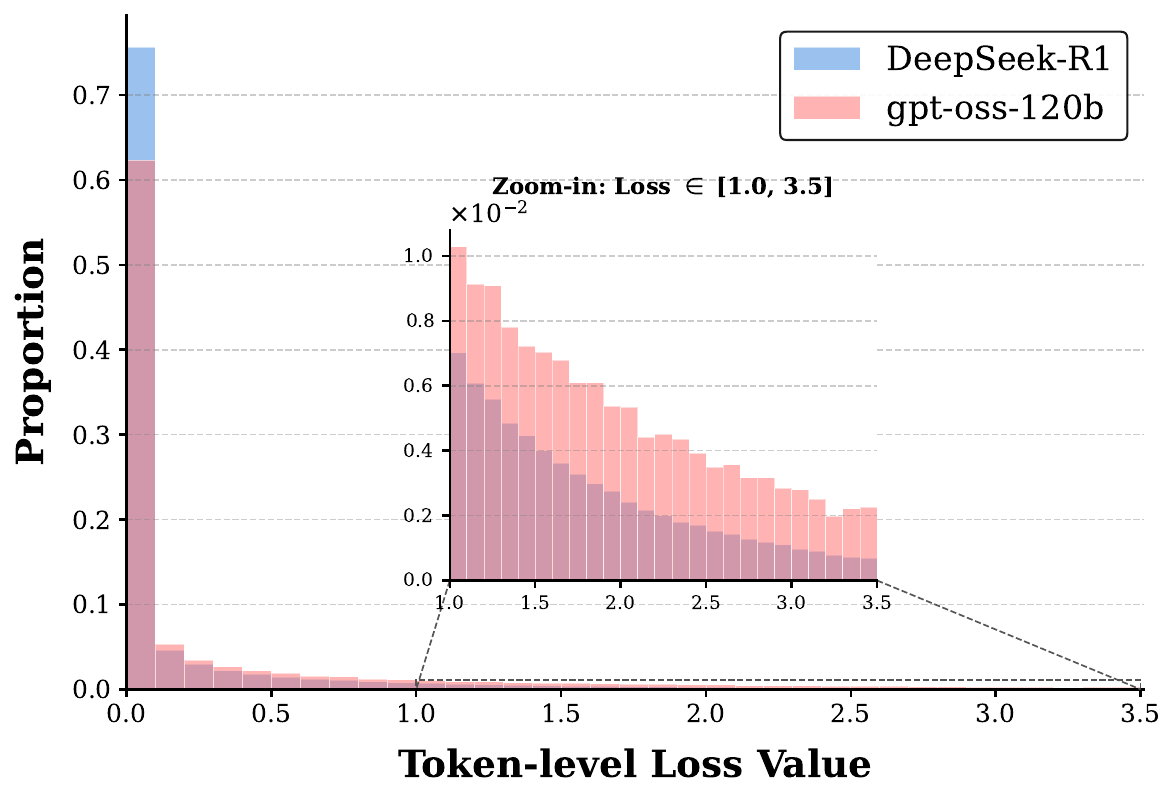}
        \caption{Qwen3-8B’s token-level loss distribution after SFT.}
    \end{subfigure}
    \begin{subfigure}[t]{0.32\textwidth}
        \centering
        \includegraphics[width=\textwidth]{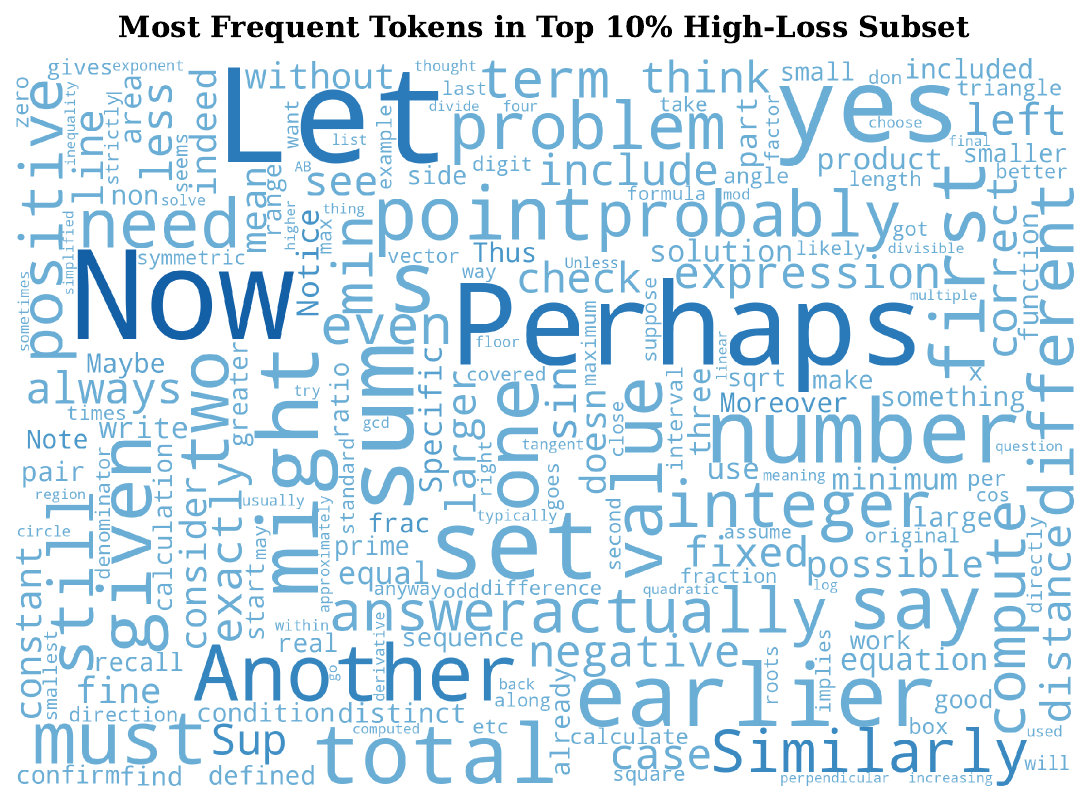}
        \caption{Most frequent tokens in the \texttt{DeepSeek-R1} data after SFT.}
    \end{subfigure}
    \begin{subfigure}[t]{0.32\textwidth}
        \centering
        \includegraphics[width=\textwidth]{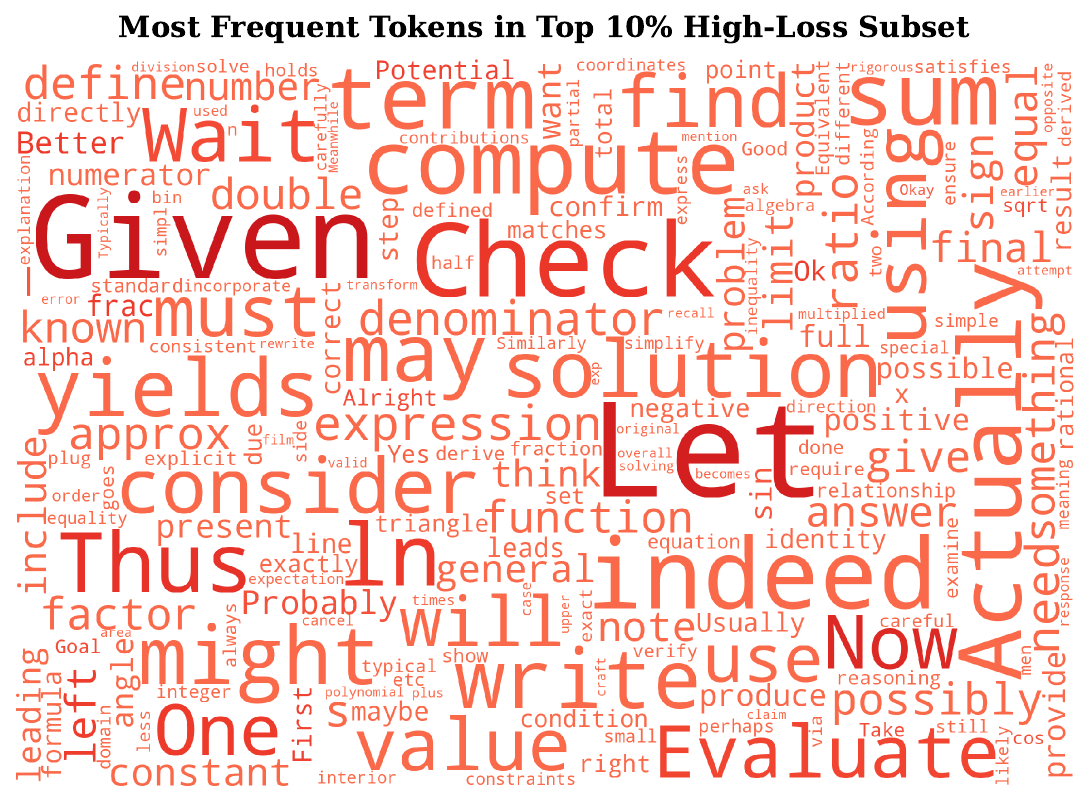}
        \caption{Most frequent tokens in the \texttt{gpt-oss-120b} data after SFT.}
    \end{subfigure}
    \caption{Token-level SFT loss analysis for the Qwen3-8B model. (a) show the token-level loss distribution after the SFT training, where \textcolor{blue}{blue}/\textcolor{red}{red} bars represent experiments with \texttt{DeepSeek-R1}/\texttt{gpt-oss-120b} trajectories. (b, c) are word clouds for the most frequent tokens in the top $10\%$ token-level loss token subset of the (\texttt{DeepSeek-R1}, \texttt{gpt-oss-120b}) data.}
    \label{fig:token-level-loss-analysis}
% \vspace{-5mm}
\end{figure}

\subsection{Token-Level SFT Loss Deconstruction and Analysis}
As a diagnostic entry-point, we calculate and visualize the token-level loss distribution for the base model before and after the SFT training with long CoT reasoning trajectories.

\textbf{The Long-Tail Distribution of Token-level SFT Loss.} 
As shown in Figures \ref{fig:token-level-loss-analysis} (a), token-level loss exhibits a severe long-tail distribution after SFT training. The vast majority of tokens ($\sim55\%$ before training (see Figure~\ref{fig:token-level-loss-analysis-qwen3-8b} in the Appendix), $\sim75\%$ after training for the \texttt{DeepSeek-R1} experiment) possess exceptionally low loss ($< 0.1$), forming the ``head'' part of the distribution. 
The remaining portion, namely the tokens exhibiting distinctly high loss, constitutes the ``long tail'' of the distribution and provides the main optimization signals during SFT. Notably, comparing the two data sources, the \texttt{gpt-oss-120b} trajectories possess a significantly thicker tail (higher proportion of high-loss tokens) compared to \texttt{DeepSeek-R1}.

\textbf{Visualization of Most Frequent Tokens in the High-loss Tail Part.}
To understand what types of token constitute the high-loss tail part, we generate word clouds to visualize the most frequent tokens in the top $10\%$ high-loss subset of all tokens (Figures \ref{fig:token-level-loss-analysis} (b) and (c), larger tokens indicate higher frequency.). 
We observe that, in both experiment groups with \texttt{DeepSeek-R1} and \texttt{gpt-oss-120b} data, the most frequent high-loss tokens are not the constituent tokens executing the internal details of individual reasoning steps; rather, they are the \textit{key reasoning tokens} which dictate logical transitions and structural shifts~\citep{qian2025demystifying,wang2025beyond} in the reasoning trajectories.
However, the semantic nature of these \textit{key reasoning tokens} diverges significantly between the two models: as summarized in Table \ref{tab:high_loss_comparison}:
for \texttt{DeepSeek-R1}, representative high-loss tokens like \texttt{`Perhaps'}, \texttt{`probably'} and \texttt{`Another'} mainly indicate a \textbf{divergent exploration} phase, where the model branches from the current reasoning path to propose a new hypothesis or alternative problem-solving angle;
for \texttt{gpt-oss-120b}, the most prominent tokens like \texttt{`Thus'}, \texttt{`indeed'}, and \texttt{`yields'} signify a \textbf{convergent exploitation} phase, where the model strengthens and pushes forward the current reasoning path.
\begin{table}[t]
% \vspace{-5mm}
    \centering
    \caption{Representative key reasoning tokens in the high-loss tail part and their average loss before/after SFT training (\texttt{\textcolor{YellowOrange}{average-loss-before-sft}} $\rightarrow$ \texttt{\textcolor{LimeGreen}{average-loss-after-sft}}).}
    \label{tab:high_loss_comparison}
    \vspace{-2mm}
    \begin{tabularx}{\textwidth}{p{2cm}| >{\centering\arraybackslash}X}
        \toprule
        \textbf{Data Source} 
            & \textbf{Representative High-Loss Tokens} \\
        \midrule
        \texttt{DeepSeek-R1} \newline
        {\small \textcolor{YellowOrange}{3.538} $\rightarrow$ \textcolor{LimeGreen}{1.533}}
            & \texttt{`Perhaps', `Now', `Another', `Let', `might', `given', `say', `Similarly', `earlier', `probably'} \\
        % \addlinespace[4pt]
        \midrule
        \texttt{gpt-oss-120b} \newline
        {\small \textcolor{YellowOrange}{4.081} $\rightarrow$ \textcolor{LimeGreen}{1.784}}
            & \texttt{`Check', `indeed', `Thus', `yields', `Given', `Let', `might', `Evaluate', `Actually', `Wait', `Now', `consider', `compute'} \\
        \bottomrule
    \end{tabularx}
\vspace{-4mm}
\end{table}
Moreover, we observe a striking \textbf{dilution effect} of SFT loss. 
As shown in Table \ref{tab:high_loss_comparison}, when we aggregate the token-level loss for these representative key reasoning tokens, we find that their average loss values are actually \textbf{quite close} (e.g., after SFT: $1.533$ for \texttt{DeepSeek-R1} vs. $1.784$ for \texttt{gpt-oss-120b}), and are significantly higher than the overall average loss. 
This reveals that the SFT loss discrepancy observed in Section \ref{sec:comparative_study} is essentially a \textit{dilution effect}: the \texttt{DeepSeek-R1} trajectories contain a lower concentration of dense reasoning transition signals, resulting in a larger proportion of low-loss ``routine'' tokens (e.g., simple calculation tokens like "3 + 2 = 5") that drag down the overall SFT loss.

\subsection{Uncovering Thought Structure Differences via Reasoning Behavior Analysis}
The token-level analysis implies that \texttt{DeepSeek-R1} and \texttt{gpt-oss-120b} utilize different structure-level thinking patterns~\citep{jiang2025makes,chen2026molecular} (i.e., "divergent exploration" versus "convergent exploitation").
To further investigate this structural difference, we decompose the macro-level continuous reasoning trajectories into micro-level discrete reasoning steps and analyze the reasoning behaviors~\citep{gandhi2025cognitive} underlying them.
Building upon~\citet{bogdan2025thought,gandhi2025cognitive}, we categorize reasoning steps into four behaviors: \texttt{\textbf{Propose}} (branch to an alternative reasoning path), \texttt{\textbf{Deduce}} (linearly push forward the current reasoning path), alongside \texttt{\textbf{Verify}} and \texttt{\textbf{Backtrack}} (which collectively serve as uncertainty management), as defined in detail in Appendix~\ref{appendix:experiment_setup_reason_behavior}, Table~\ref{tab:reasoning_behavior_intro}. We utilize a powerful open-weight LLM, \texttt{DeepSeek-V3.2}~\citep{deepseekai2025deepseekv32pushingfrontieropen}, with a clearly-defined and well-designed prompt template to automatically annotate the reasoning steps (see Appendix~\ref{appendix:experiment_setup_reason_behavior} for the details of the annotation experiment).

\begin{figure}[t]
    \centering
    \captionsetup[subfigure]{justification=centering} 
    
    % 第一行
    \begin{subfigure}[t]{0.32\textwidth}
        \centering
        \includegraphics[width=\textwidth]{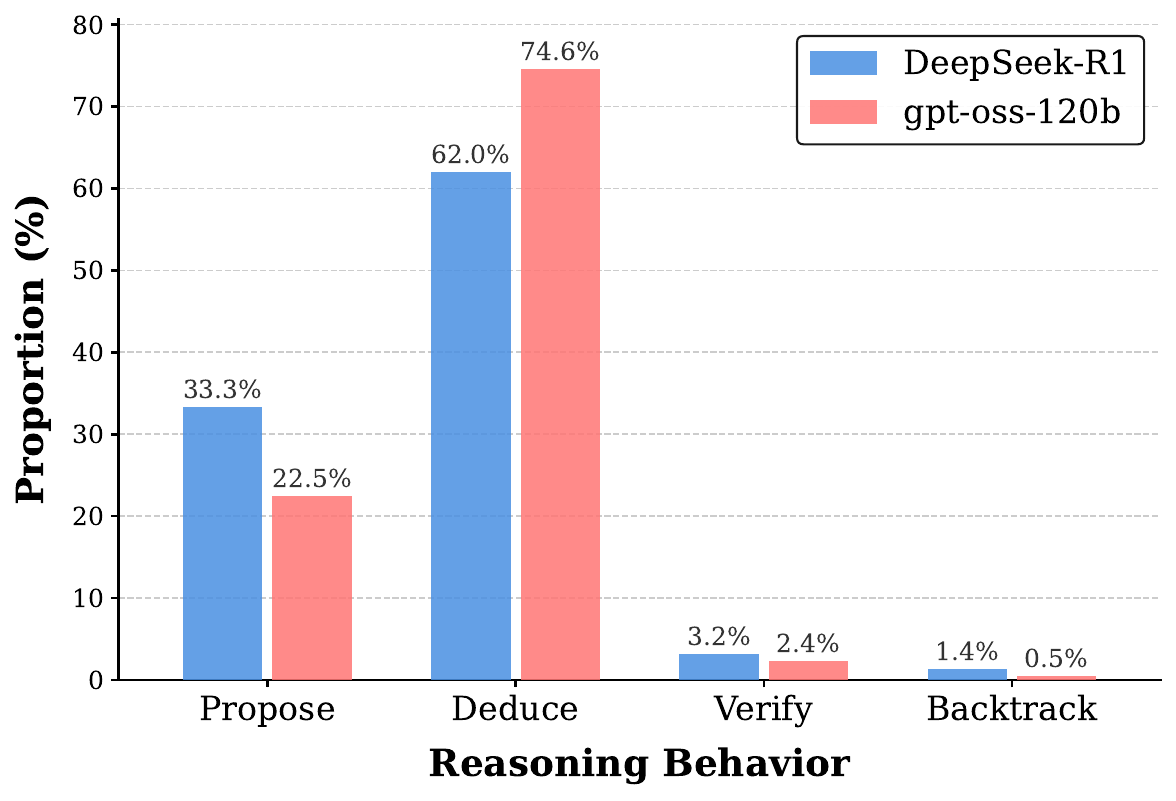}
        \caption{Distribution: training data.}
    \end{subfigure}% % <-- 注意这个百分号，防止产生多余空格
    % \hfill % <-- 自动撑开水平间距，让左右两图对齐更美观
    \begin{subfigure}[t]{0.65\textwidth}
        \centering
        \includegraphics[width=\textwidth]{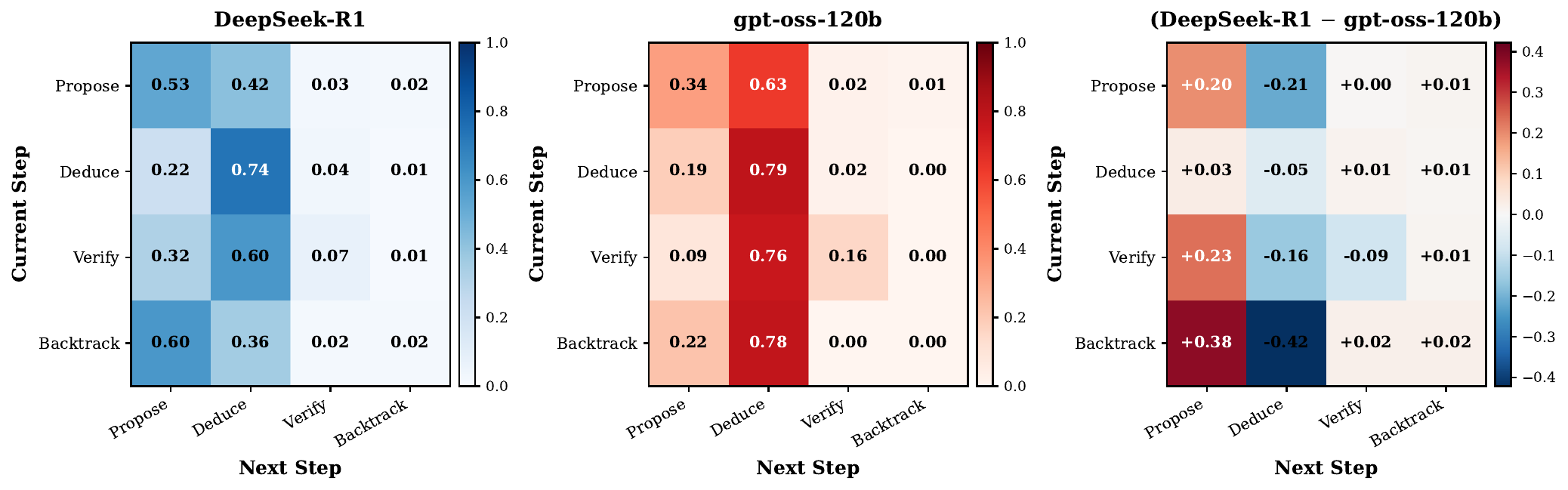}
        \caption{Transition matrix for training data.} % <-- 稍微精简了这句的冗长，保证它和左边大概率在一行内
    \end{subfigure}
    
    % \vspace{-1mm} % �� 核心秘籍 1：用负间距强制把第二行往上拽！(如果还是松，可以改成 -5mm)
    
    % 第二行
    \begin{subfigure}[t]{0.32\textwidth}
        \centering
        \includegraphics[width=\textwidth]{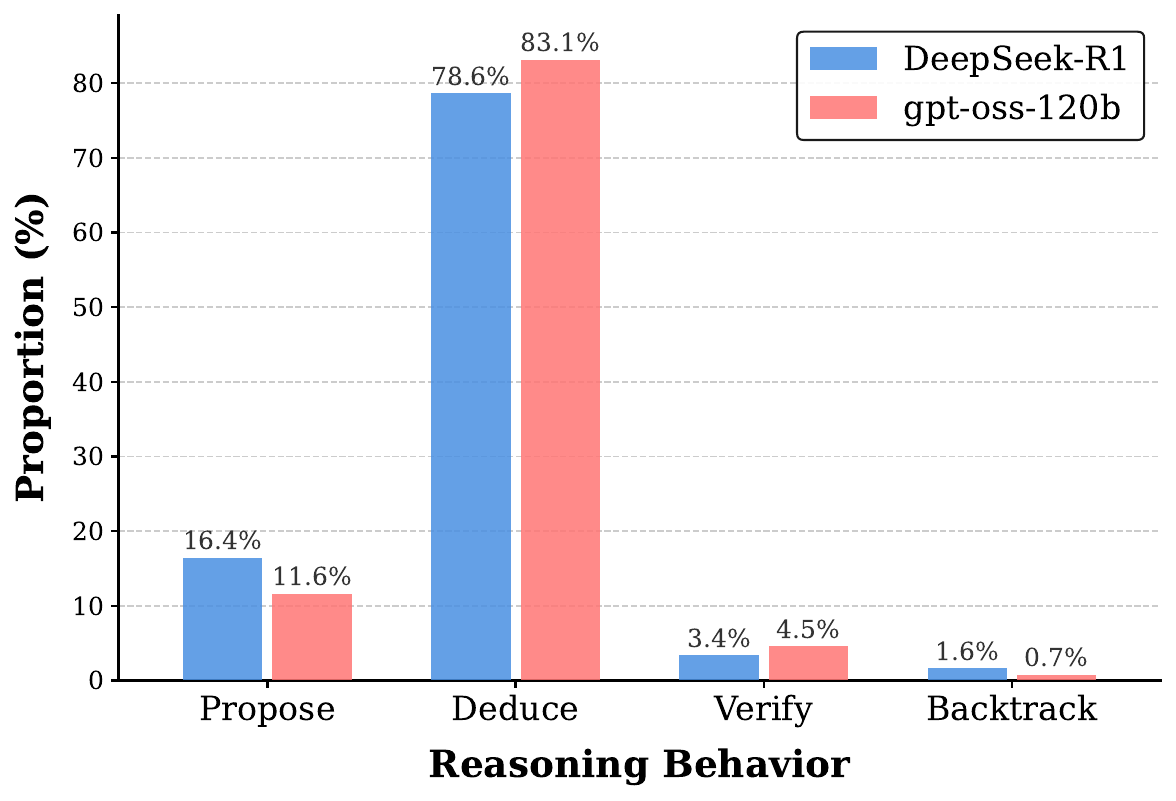}
        \caption{distribution: generated solutions for AIME24.}
    \end{subfigure}%
    % \hfill
    \begin{subfigure}[t]{0.65\textwidth}
        \centering
        \includegraphics[width=\textwidth]{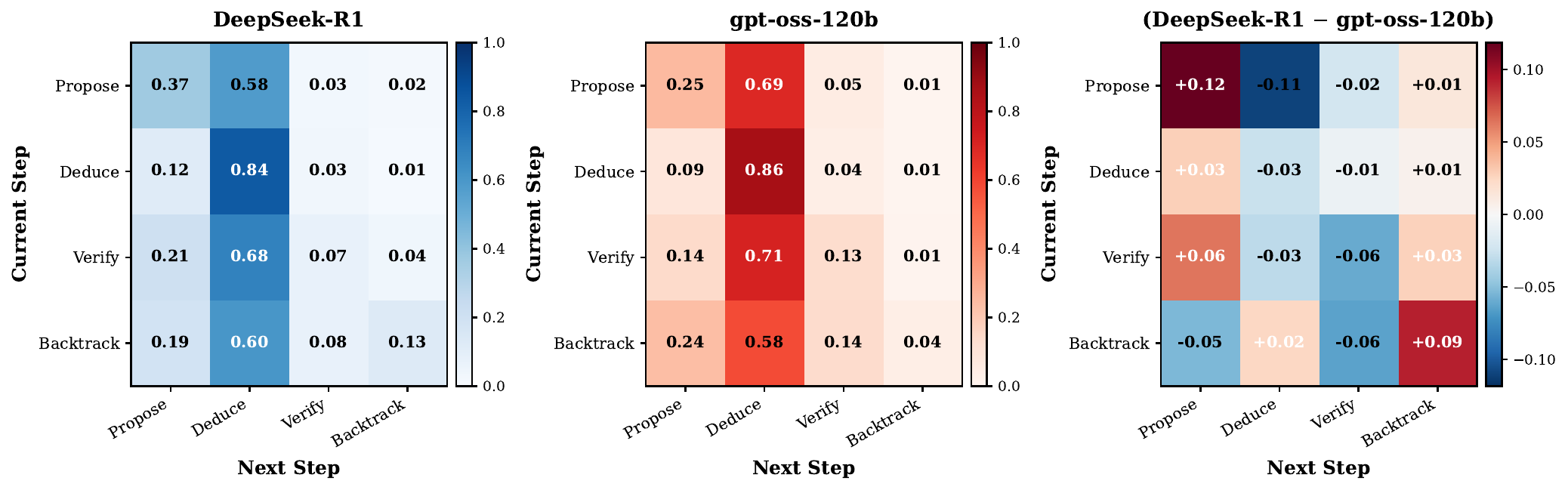}
        \caption{Transition matrix for generated AIME24 solutions.}
    \end{subfigure}
    
    \vspace{-2mm} % �� 核心秘籍 2：稍微拉近子图和总Caption的距离
    \caption{Reasoning behavior distributions (a, c) and transition matrices (b, d) for reasoning trajectories used for training and generated for solving AIME24 problems with fine-tuned Qwen3-8B. \textcolor{blue}{Blue}/\textcolor{red}{red} objects: experiments with \texttt{DeepSeek-R1}/\texttt{gpt-oss-120b}-generated data.}
    \label{fig:reason_behavior_distribution_main_text}
    \vspace{-6mm} % �� 核心秘籍 3：极其关键！压缩整个Figure和下方正文的间距，榨干最后一滴页面空间！
\end{figure}

\textbf{Behavior Distribution and Markov Transition Matrix.}
Figure \ref{fig:reason_behavior_distribution_main_text} (a) compares the behavior distribution in the original reasoning trajectories generated by \texttt{DeepSeek-R1} (\textcolor{blue}{blue} bars) and \texttt{gpt-oss-120b} (\textcolor{red}{red} bars) that are used for SFT training. Quantitatively, we observe that \texttt{DeepSeek-R1} trajectories contain a significantly higher proportion of \texttt{Propose} steps ($33.3\%$) compared to \texttt{gpt-oss-120b} ($22.5\%$). 
In contrast, the proportion of \texttt{Deduce} steps in \texttt{DeepSeek-R1} ($62.0\%$) is markedly lower than that in \texttt{gpt-oss-120b} ($74.6\%$). Furthermore, the Markov transition matrices~\citep{chen2026molecular} of the reasoning behavior (Figure \ref{fig:reason_behavior_distribution_main_text} (b)) reveal distinct step-to-step behavior transition patterns:  \texttt{DeepSeek-R1} exhibits notably higher probabilities of transitioning into the \texttt{Propose} state. Notably, it displays a more frequent \texttt{Propose} $\rightarrow$ \texttt{Propose} transition pattern (transition probability$=0.53$) in comparison with \texttt{gpt-oss-120b} (transition probability$=0.34$), indicating that \texttt{DeepSeek-R1} is more inclined to generate \textbf{consecutive reasoning branches} before committing to a specific deductive chain.

According to the behavioral definitions, \texttt{Propose} signifies exploring a new idea or branching to an alternative path, while \texttt{Deduce} represents executing linear, forward logical inferences. 
The above statistics reveal that \texttt{DeepSeek-R1} inherently favors a highly exploratory reasoning pattern characterized by frequently branching and pivoting. 
In contrast, \texttt{gpt-oss-120b} strongly leans toward convergent exploitation, focusing on deep, continuous deductive chains. 
The frequent \texttt{Propose} $\rightarrow$ \texttt{Propose} transitions in \texttt{DeepSeek-R1} suggest that \textit{\textbf{the model often engages in continuous branching without executing deep deductive analysis on the newly proposed ideas}} (see Appendix~\ref{appendix:long_cot_showcase}, Figure~\ref{figure:showcase_training_data} for a specific example). 
This structural bias implies a high degree of \textit{redundancy} within the \texttt{DeepSeek-R1} trajectories, where many shallow exploratory paths may not meaningfully contribute to the final logical resolution.
Crucially, this thought structure is directly transferred to the student models during SFT. 
As shown in Figures \ref{fig:reason_behavior_distribution_main_text} (c) and (d), the Qwen3-8B model fine-tuned on \texttt{DeepSeek-R1} data generates AIME24 solutions that inherit this exploratory distribution, exhibiting a $12\%$ higher probability of \texttt{Propose} $\rightarrow$ \texttt{Propose} transitions compared to its \texttt{gpt-oss-120b}-trained counterpart. 
Consequently, the student model wastes extensive context on shallow branching, and sometimes iteratively re-proposes previously explored dead-ends (see a specific example in Appendix~\ref{appendix:long_cot_showcase}, Figure~\ref{figure:showcase_testing_data}). 
This inherited inefficiency \textbf{\textit{hinders the student model's ability to converge on the correct answer, explaining its degraded generalization performance}}.

\textbf{Validating Exploratory Redundancy via Random Reasoning Step Deletion.}
\begin{figure}[t]
    % \vspace{-4mm}
    \centering
    \captionsetup[subfigure]{justification=centering} 
    % 第一行
    \begin{subfigure}[t]{0.35\textwidth}
        \centering
        \includegraphics[width=\textwidth]{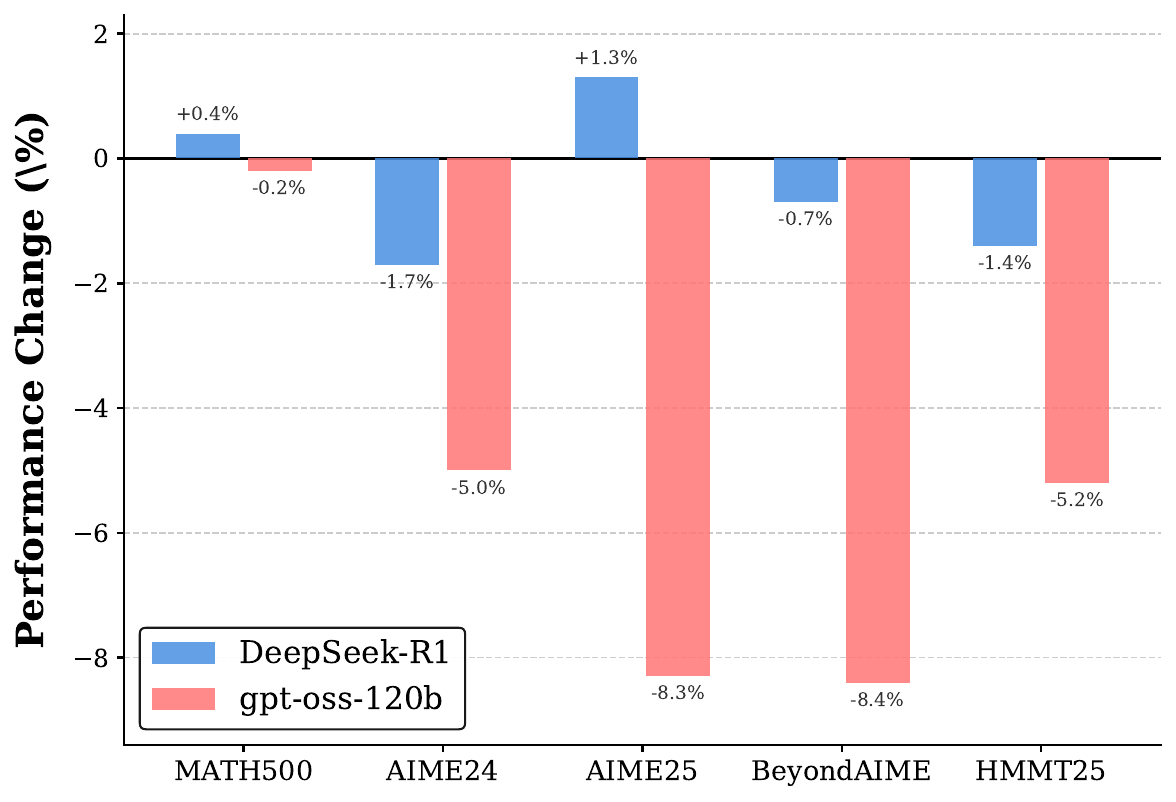}
        \caption{Qwen3-8B results.}
    \end{subfigure}
    \hspace{0.06\textwidth}   % ← 加在这里，数值可按需调整
    \begin{subfigure}[t]{0.35\textwidth}
        \centering
        \includegraphics[width=\textwidth]{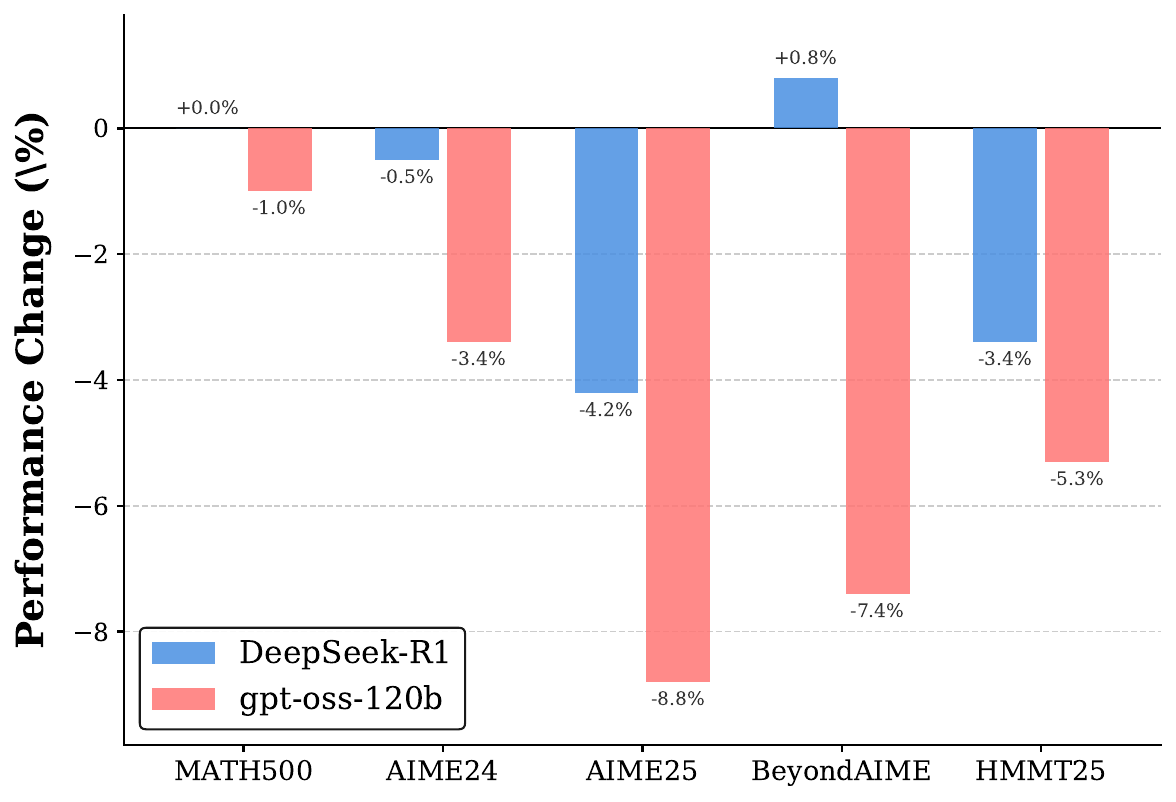}
        \caption{Qwen2.5-32B results.}
    \end{subfigure}
    \vspace{-2mm}
    \caption{Performance change ratio ($(\text{Acc}_{\text{original}}-\text{Acc}_{\text{retrain}})/\text{Acc}_{\text{original}}$) on five benchmarks after randomly deleting 10\% reasoning steps in each training trajectory. \textcolor{blue}{Blue}/\textcolor{red}{red} bars represent experiments with \texttt{DeepSeek-R1}/\texttt{gpt-oss-120b}-generated data, respectively.}
    \label{fig:random_del_10p_reason_step_main_context}
    \vspace{-4mm}
\end{figure}
To empirically validate our hypothesis that \texttt{DeepSeek-R1} trajectories contain a higher degree of structural redundancy, we design a comparison experiment: for each training trajectory in both datasets, we \textbf{randomly} delete $10\%$ of its reasoning steps and retrain the base models from scratch. 
If a trajectory is a dense, highly-dependent deductive chain, the deletion should break the logical flow and degrade performance more drastically. 
Conversely, if a trajectory has many redundant branches, the deletion should have a minor impact. 
The results in Figure \ref{fig:random_del_10p_reason_step_main_context} confirm this hypothesis. Models trained on the $10\%$-deleted \texttt{gpt-oss-120b} data suffer more significant performance drops. In contrast, models trained on the $10\%$-deleted \texttt{DeepSeek-R1} data exhibit minimal degradation. Surprisingly, in several instances (e.g., Qwen3-8B on MATH500 and AIME25; Qwen2.5-32B on BeyondAIME), the performance actually \textit{increases}. This suggests that the \texttt{DeepSeek-R1} data inherently contain more redundant sub-paths. In fact, moderately truncating these exploratory steps may act as an implicit regularization~\citep{chen-etal-2024-masked}, preventing the student model from overfitting to superficial trial-and-error formatting and forcing it to focus on the core deductive reasoning path.
\begin{insightbox}
\textbf{Takeaway:}
\texttt{DeepSeek-R1} trajectories favor a pattern of frequently branching into alternative ideas without deep deduction, introducing more redundant reasoning paths, which is inherited by student models during SFT and potentially degrades their generalization.
\end{insightbox}

\section{Improving SFT by Filtering out Frequently Branching Trajectories}
\begin{table}[b]
\vspace{-4mm}
\centering
\caption{SFT performance after removing trajectories with top difference ratio of reasoning step number (proxy metric 1) from the \texttt{DeepSeek-R1} generated dataset.}
\label{tab:rm_top_diff}
\vspace{-2mm}
\resizebox{\textwidth}{!}{%
\begin{tabular}{@{}l|c|ccccc|c@{}}
\toprule
\textbf{Base Model} & \textbf{Data Source} & \textbf{MATH500} & \textbf{AIME24} & \textbf{AIME25} & \textbf{BeyondAIME} & \textbf{HMMT25} & \textbf{Avg} \\
\midrule
\multirow{3}{*}{\shortstack{Qwen3-8B \\ {\footnotesize (context len: 32K)}}} 
 & full \texttt{DeepSeek-R1} set & $96.8\%$ & $63.0\%$ & $54.0\%$ & $24.5\%$ & $48.5\%$ & $57.4\%$ \\
 \cmidrule{2-8}
 & removing top $10\%$ & $97.4\%$ {\footnotesize \textcolor{green!60!black}{(\textbf{+0.6\%})}} & $64.7\%${\footnotesize \textcolor{green!60!black}{(\textbf{+1.7\%})}} & $55.7\%${\footnotesize \textcolor{green!60!black}{(\textbf{+1.7\%})}} & $25.1\%${\footnotesize \textcolor{green!60!black}{(\textbf{+0.6\%})}} & $51.5\%${\footnotesize \textcolor{green!60!black}{(\textbf{+3.0\%})}} & $\mathbf{58.9\%}$ {\footnotesize \textcolor{green!60!black}{(\textbf{+1.5\%})}} \\
 & removing top $20\%$ & $97.2\%$ {\footnotesize \textcolor{green!60!black}{(\textbf{+0.4\%})}}& $64.5\%$ {\footnotesize \textcolor{green!60!black}{(\textbf{+1.5\%})}}& $54.7\%$ {\footnotesize \textcolor{green!60!black}{(\textbf{+0.7\%})}}& $25.7\%$ {\footnotesize \textcolor{green!60!black}{(\textbf{+1.2\%})}} & $50.1\%$ {\footnotesize \textcolor{green!60!black}{(\textbf{+1.6\%})}}& $\mathbf{58.4\%}$ {\footnotesize \textcolor{green!60!black}{(\textbf{+1.0\%})}} \\
\midrule
\midrule
\multirow{3}{*}{\shortstack{Qwen2.5-7B \\ {\footnotesize (context len: 32K)}}} 
 & full \texttt{DeepSeek-R1} set & $95.0\%$ & $52.1\%$ & $44.8\%$ & $16.7\%$ & $38.7\%$ & $49.5\%$ \\
  \cmidrule{2-8}
 & removing top $10\%$& $95.0\%$ & $54.3\%${\footnotesize \textcolor{green!60!black}{(\textbf{+2.2\%})}} & $46.7\%${\footnotesize \textcolor{green!60!black}{(\textbf{+1.9\%})}} & $19.3\%${\footnotesize \textcolor{green!60!black}{(\textbf{+2.6\%})}} & $40.2\%${\footnotesize \textcolor{green!60!black}{(\textbf{+1.5\%})}} & $\mathbf{51.1\%}$ {\footnotesize \textcolor{green!60!black}{(\textbf{+1.6\%})}}\\
 & removing top $20\%$& $95.8\%${\footnotesize \textcolor{green!60!black}{(\textbf{+0.8\%})}} & $52.2\%${\footnotesize \textcolor{green!60!black}{(\textbf{+0.1\%})}} & $44.4\%${\footnotesize \textcolor{red!60!black}{(\textbf{-0.4\%})}} & $16.2\%${\footnotesize \textcolor{red!60!black}{(\textbf{-0.5\%})}} & $40.0\%${\footnotesize \textcolor{green!60!black}{(\textbf{+1.3\%})}} & $\mathbf{49.7\%}$ {\footnotesize \textcolor{green!60!black}{(\textbf{+0.2\%})}}\\
 \midrule
\midrule
\multirow{3}{*}{\shortstack{Qwen2.5-7B \\ {\footnotesize (context len: 96K)}}} 
  & full \texttt{DeepSeek-R1} set  & $93.6\%$ & $61.7\%$ & $53.4\%$ & $27.5\%$ & $38.3\%$ & $54.9\%$ \\
 \cmidrule{2-8}
 & removing top $10\%$& $95.8\%${\footnotesize \textcolor{green!60!black}{(\textbf{+2.2\%})}} & $65.6\%$ {\footnotesize \textcolor{green!60!black}{(\textbf{+3.9\%})}}& $55.2\%$ {\footnotesize \textcolor{green!60!black}{(\textbf{+1.8\%})}}& $30.3\%$ {\footnotesize \textcolor{green!60!black}{(\textbf{+2.8\%})}}& $42.4\%$ {\footnotesize \textcolor{green!60!black}{(\textbf{+4.1\%})}}& $\mathbf{57.9\%}$ {\footnotesize \textcolor{green!60!black}{(\textbf{+3.0\%})}}\\
 & removing top $20\%$& $95.6\%${\footnotesize \textcolor{green!60!black}{(\textbf{+2.0\%})}} & $64.8\%${\footnotesize \textcolor{green!60!black}{(\textbf{+3.1\%})}} & $53.1\%${\footnotesize \textcolor{red!60!black}{(\textbf{-0.3\%})}} & $29.2\%${\footnotesize \textcolor{green!60!black}{(\textbf{+1.7\%})}} & $39.3\%${\footnotesize \textcolor{green!60!black}{(\textbf{+1.0\%})}} & $\mathbf{56.4\%}$ {\footnotesize \textcolor{green!60!black}{(\textbf{+1.5\%})}}\\
\bottomrule
\end{tabular}%
}
\end{table}
In Section \ref{sec:mechanistic_analysis}, we show that \texttt{DeepSeek-R1} exhibits a reasoning pattern heavily biased toward divergent exploration, frequently branching into alternative ideas without executing deep deduction. When SFT on these long CoT trajectories, the base models internalize this pattern and get trapped in redundant exploratory branches, leading to inferior generalization performance on reasoning benchmarks (Section~\ref{sec:comparative_study}). 
To further validate this insight and provide a practical recipe for enhancing Long CoT SFT, we propose a targeted data intervention: \textbf{\textit{filtering out the most frequently branching trajectories from the original \texttt{DeepSeek-R1} dataset}}, which may mitigate the base model's tendency to internalize this reasoning pattern. 
Ideally, identifying such trajectories would involve annotating every reasoning step across the 500K reasoning trajectories with LLMs. However, executing this at scale is computationally prohibitive: there are over 200 steps in each trajectory on average. 
Therefore, we design two \textit{proxy metrics} to identify the frequently branching trajectories.

\textbf{Proxy 1: Difference Ratio of Reasoning Step Number.}
Our first proxy leverages the paired nature of our datasets. For the exact same prompt, we compute the \textit{difference in the number of reasoning steps between the \texttt{DeepSeek-R1} and \texttt{gpt-oss-120b} trajectories, normalized by the total steps of the \texttt{DeepSeek-R1} trajectory}. 
Intuitively, a higher difference ratio indicates that the \texttt{DeepSeek-R1} trajectory contains significantly more redundant exploration steps compared to its more deductive \texttt{gpt-oss-120b} counterpart. 
Based on this metric, we remove the trajectories with the highest difference ratios (top 10\% and top 20\%) and retrain the models, ensuring that all other settings and total training steps remain identical to the original SFT experiments. As shown in Table \ref{tab:rm_top_diff}, this simple data filtering strategy yields consistent performance gains. For instance, by removing the $10\%$ trajectories, the average accuracy on five benchmarks of Qwen3-8B and Qwen2.5-7B improves by $1.5\%$ and $3.0\%$, respectively.

\begin{table}[t]
\centering
\caption{
SFT performance using the bottom/top K\% subset (in terms of the proportion of the reasoning steps containing “branching”-related keywords, proxy metric 2) of trajectories.
}
\label{tab:bottom_select}
%\vspace{-2mm}
\resizebox{\textwidth}{!}{%
\begin{tabular}{@{}l|c|ccccc|c@{}}
\toprule
\textbf{Base Model} & \textbf{Data Source} & \textbf{MATH500} & \textbf{AIME24} & \textbf{AIME25} & \textbf{BeyondAIME} & \textbf{HMMT25} & \textbf{Avg} \\
\midrule
\multirow{7}{*}{\shortstack{Qwen3-8B \\ {\footnotesize (context len: 32K)}}} 
 & full \texttt{DeepSeek-R1} set (original SFT) & $96.8\%$\scriptsize$\pm0.02\%$ & $63.0\%$\scriptsize$\pm0.03\%$ & $54.0\%$\scriptsize$\pm0.03\%$ & $24.5\%$\scriptsize$\pm0.03\%$ & $48.5\%$\scriptsize$\pm0.03\%$ & $57.4\%$ \\
 \cmidrule{2-8}
 & \texttt{bottom $\mathbf{90\%}$ (experiment group)} & \cellcolor{green!25}$97.0\%$\scriptsize$\pm0.02\%$ & \cellcolor{green!25}$64.7\%$\scriptsize$\pm0.03\%$ & \cellcolor{green!25}$55.5\%$\scriptsize$\pm0.03\%$ & \cellcolor{green!25}$24.9\%$\scriptsize$\pm0.03\%$ & \cellcolor{green!25}$48.8\%$\scriptsize$\pm0.03\%$ & $\mathbf{58.2\%}$ {\footnotesize \textcolor{green!60!black}{(\textbf{+0.8\%})}} \\
  & \texttt{top $\mathbf{90\%}$ (control group)} & \cellcolor{green!25}$97.0\%$\scriptsize$\pm0.02\%$ & $63.0\%$\scriptsize$\pm0.03\%$ & \cellcolor{red!25}$52.3\%$\scriptsize$\pm0.03\%$ & \cellcolor{red!25}$23.9\%$\scriptsize$\pm0.03\%$ & \cellcolor{red!25}$47.7\%$\scriptsize$\pm0.03\%$ & $56.8\%$ {\footnotesize \textcolor{red!60!black}{(\textbf{-0.6\%})}} \\
\cmidrule{2-8}
 & \texttt{bottom $\mathbf{80\%}$ (experiment group)} & \cellcolor{green!25}$97.0\%$\scriptsize$\pm0.02\%$ & \cellcolor{green!25}$64.7\%$\scriptsize$\pm0.03\%$ & \cellcolor{green!25}$55.7\%$\scriptsize$\pm0.03\%$ & \cellcolor{green!25}$27.1\%$\scriptsize$\pm0.03\%$ & \cellcolor{green!25}$52.3\%$\scriptsize$\pm0.03\%$ & $\mathbf{59.4\%}$ {\footnotesize \textcolor{green!60!black}{(\textbf{+2.0\%})}} \\
  & \texttt{top $\mathbf{80\%}$ (control group)} & $96.8\%$\scriptsize$\pm0.02\%$ & \cellcolor{green!25}$63.3\%$\scriptsize$\pm0.03\%$ & \cellcolor{red!25}$52.1\%$\scriptsize$\pm0.03\%$ & \cellcolor{red!25}$22.5\%$\scriptsize$\pm0.02\%$ & \cellcolor{red!25}$47.4\%$\scriptsize$\pm0.03\%$ & $56.4\%$ {\footnotesize \textcolor{red!60!black}{(\textbf{-1.0\%})}} \\
\cmidrule{2-8}
 & \texttt{bottom $\mathbf{50\%}$ (experiment group)} & \cellcolor{green!25}$97.0\%$\scriptsize$\pm0.02\%$ & \cellcolor{green!25}$66.5\%$\scriptsize$\pm0.03\%$ & \cellcolor{green!25}$59.1\%$\scriptsize$\pm0.03\%$ & \cellcolor{green!25}$25.1\%$\scriptsize$\pm0.03\%$ & \cellcolor{green!25}$50.5\%$\scriptsize$\pm0.03\%$ & $\mathbf{59.6\%}$ {\footnotesize \textcolor{green!60!black}{(\textbf{+2.2\%})}} \\
  & \texttt{top $\mathbf{50\%}$ (control group)} & \cellcolor{red!25}$96.0\%$\scriptsize$\pm0.02\%$ & $63.0\%$\scriptsize$\pm0.03\%$ & \cellcolor{red!25}$52.3\%$\scriptsize$\pm0.03\%$ & \cellcolor{red!25}$22.7\%$\scriptsize$\pm0.02\%$ & \cellcolor{red!25}$47.1\%$ \scriptsize$\pm0.03\%$& $56.2\%$ {\footnotesize \textcolor{red!60!black}{(\textbf{-1.2\%})}} \\
\midrule
\midrule
\multirow{7}{*}{\shortstack{Qwen2.5-7B \\ {\footnotesize (context len: 32K)}}} 
 & full \texttt{DeepSeek-R1} set (original SFT) & $95.0\%$\scriptsize$\pm0.02\%$ & $52.1\%$\scriptsize$\pm0.03\%$ & $44.8\%$\scriptsize$\pm0.03\%$ & $16.7\%$\scriptsize$\pm0.02\%$ & $38.7\%$\scriptsize$\pm0.03\%$ & $49.5\%$ \\
   \cmidrule{2-8}
 & \texttt{bottom $\mathbf{90\%}$ (experiment group)} & \cellcolor{green!25}$95.4\%$\scriptsize$\pm0.02\%$ & \cellcolor{green!25}$53.7\%$\scriptsize$\pm0.03\%$ & \cellcolor{green!25}$46.5\%$\scriptsize$\pm0.03\%$ & \cellcolor{green!25}$19.3\%$\scriptsize$\pm0.03\%$ & \cellcolor{green!25}$41.8\%$\scriptsize$\pm0.03\%$ & $\mathbf{51.3\%}$ {\footnotesize \textcolor{green!60!black}{(\textbf{+1.8\%})}} \\
  & \texttt{top $\mathbf{90\%}$ (control group)}& $95.0\%$\scriptsize$\pm0.02\%$ & \cellcolor{red!25}$51.9\%$\scriptsize$\pm0.03\%$ & \cellcolor{green!25}$44.9\%$\scriptsize$\pm0.03\%$ & \cellcolor{green!25}$17.7\%$\scriptsize$\pm0.02\%$ & \cellcolor{red!25}$37.1\%$\scriptsize$\pm0.03\%$ & $49.3\%$ {\footnotesize \textcolor{red!60!black}{(\textbf{-0.2\%})}} \\
\cmidrule{2-8}
 & \texttt{bottom $\mathbf{80\%}$ (experiment group)} & \cellcolor{red!25}$94.6\%$\scriptsize$\pm0.02\%$ & \cellcolor{green!25}$55.5\%$\scriptsize$\pm0.03\%$ & \cellcolor{green!25}$47.2\%$\scriptsize$\pm0.03\%$ & \cellcolor{green!25}$18.5\%$\scriptsize$\pm0.02\%$ & \cellcolor{green!25}$40.1\%$\scriptsize$\pm0.03\%$ & $\mathbf{51.2\%}$ {\footnotesize \textcolor{green!60!black}{(\textbf{+1.7\%})}} \\
  & \texttt{top $\mathbf{80\%}$ (control group)} & \cellcolor{red!25}$94.4\%$\scriptsize$\pm0.02\%$ & \cellcolor{red!25}$52.0\%$\scriptsize$\pm0.03\%$ & \cellcolor{red!25}$44.5\%$\scriptsize$\pm0.03\%$ & \cellcolor{red!25}$15.0\%$\scriptsize$\pm0.02\%$& \cellcolor{green!25}$39.3\%$\scriptsize$\pm0.03\%$ & $49.0\%$ {\footnotesize \textcolor{red!60!black}{(\textbf{-0.5\%})}} \\
 \cmidrule{2-8}
 & \texttt{bottom $\mathbf{50\%}$ (experiment group)}  & \cellcolor{green!25}$95.4\%$\scriptsize$\pm0.02\%$ & \cellcolor{red!25}$51.9\%$\scriptsize$\pm0.03\%$ & \cellcolor{green!25}$46.0\%$\scriptsize$\pm0.03\%$ & \cellcolor{green!25}$19.1\%$\scriptsize$\pm0.02\%$ & \cellcolor{green!25}$41.0\%$\scriptsize$\pm0.03\%$ & $\mathbf{50.7\%}$ {\footnotesize \textcolor{green!60!black}{(\textbf{+1.2\%})}} \\
  & \texttt{top $\mathbf{50\%}$ (control group)} & \cellcolor{red!25}$94.6\%$\scriptsize$\pm0.02\%$ & \cellcolor{red!25}$50.6\%$\scriptsize$\pm0.03\%$ & \cellcolor{red!25}$43.2\%$\scriptsize$\pm0.03\%$ & \cellcolor{red!25}$15.0\%$\scriptsize$\pm0.02\%$ & \cellcolor{red!25}$38.3\%$\scriptsize$\pm0.03\%$ & $48.3\%$ {\footnotesize \textcolor{red!60!black}{(\textbf{-1.2\%})}} \\
  \midrule
  \midrule
\multirow{7}{*}{\shortstack{Qwen2.5-7B \\ {\footnotesize (context len: 96K)}}} 
 & full \texttt{DeepSeek-R1} set (original SFT)  & $93.6\%$\scriptsize$\pm0.02\%$ & $61.7\%$\scriptsize$\pm0.03\%$ & $53.4\%$\scriptsize$\pm0.03\%$ & $27.5\%$\scriptsize$\pm0.02\%$ & $38.3\%$\scriptsize$\pm0.03\%$ & $54.9\%$ \\
  \cmidrule{2-8}
 & \texttt{bottom $\mathbf{90\%}$ (experiment group)} & \cellcolor{green!25}$95.2\%$\scriptsize$\pm0.02\%$ & \cellcolor{green!25}$64.9\%$\scriptsize$\pm0.03\%$ & \cellcolor{green!25}$57.2\%$\scriptsize$\pm0.03\%$ & \cellcolor{green!25}$33.0\%$\scriptsize$\pm0.03\%$ & \cellcolor{green!25}$42.1\%$\scriptsize$\pm0.03\%$ & $\mathbf{58.5\%}$ {\footnotesize \textcolor{green!60!black}{(\textbf{+3.6\%})}} \\
  & \texttt{top $\mathbf{90\%}$ (control group)}& \cellcolor{red!25}$93.4\%$\scriptsize$\pm0.02\%$ & \cellcolor{red!25}$61.6\%$\scriptsize$\pm0.03\%$ & \cellcolor{green!25}$54.0\%$\scriptsize$\pm0.03\%$ & \cellcolor{red!25}$26.6\%$\scriptsize$\pm0.03\%$ & \cellcolor{red!25}$37.1\%$\scriptsize$\pm0.03\%$ & $54.5\%$ {\footnotesize \textcolor{red!60!black}{(\textbf{-0.4\%})}} \\
\cmidrule{2-8}
 & \texttt{bottom $\mathbf{80\%}$ (experiment group)} & \cellcolor{green!25}$95.8\%$\scriptsize$\pm0.02\%$ & \cellcolor{green!25}$65.0\%$\scriptsize$\pm0.03\%$ & \cellcolor{green!25}$56.8\%$\scriptsize$\pm0.03\%$ & \cellcolor{green!25}$32.1\%$\scriptsize$\pm0.03\%$ & \cellcolor{green!25}$40.5\%$\scriptsize$\pm0.03\%$ & $\mathbf{58.0\%}$ {\footnotesize \textcolor{green!60!black}{(\textbf{+3.1\%})}} \\
  & \texttt{top $\mathbf{80\%}$ (control group)} & \cellcolor{red!25}$93.2\%$\scriptsize$\pm0.02\%$ & \cellcolor{red!25}$60.8\%$\scriptsize$\pm0.03\%$ & \cellcolor{red!25}$52.4\%$\scriptsize$\pm0.03\%$ & \cellcolor{red!25}$27.0\%$\scriptsize$\pm0.03\%$ & \cellcolor{green!25}$39.3\%$\scriptsize$\pm0.03\%$ & $54.5\%$ {\footnotesize \textcolor{red!60!black}{(\textbf{-0.4\%})}} \\
 \cmidrule{2-8}
 & \texttt{bottom $\mathbf{50\%}$ (experiment group)}  & \cellcolor{green!25}$94.4\%$\scriptsize$\pm0.02\%$ & \cellcolor{red!25}$61.0\%$\scriptsize$\pm0.03\%$ & \cellcolor{green!25}$54.1\%$\scriptsize$\pm0.03\%$ & \cellcolor{green!25}$29.2\%$\scriptsize$\pm0.03\%$ & \cellcolor{green!25}$40.5\%$\scriptsize$\pm0.03\%$ & $\mathbf{55.8\%}$ {\footnotesize \textcolor{green!60!black}{(\textbf{+0.9\%})}} \\
  & \texttt{top $\mathbf{50\%}$ (control group)} & \cellcolor{red!25}$90.4\%$\scriptsize$\pm0.02\%$ & \cellcolor{red!25}$54.0\%$\scriptsize$\pm0.03\%$ & \cellcolor{red!25}$49.9\%$\scriptsize$\pm0.03\%$ & \cellcolor{red!25}$23.9\%$\scriptsize$\pm0.03\%$ & \cellcolor{red!25}$34.5\%$\scriptsize$\pm0.03\%$ & $50.5\%$ {\footnotesize \textcolor{red!60!black}{(\textbf{-4.4\%})}} \\
\bottomrule
\end{tabular}
}
\vspace{-4mm}
\end{table}

\textbf{Proxy 2: Proportion of Steps Containing Branching Keywords.}
Proxy 1 requires paired data from \texttt{DeepSeek-R1} and \texttt{gpt-oss-120b}. 
To establish a standalone filtering criterion, we introduce proxy 2, in which we calculate the proportion of reasoning steps within each \texttt{DeepSeek-R1} trajectory that contain explicit branching signal tokens (e.g., \texttt{"Perhaps"}, \texttt{"Another"}, \texttt{"Alternatively"}). 
Building upon our token-level findings in Section \ref{sec:mechanistic_analysis}, trajectories with a high density of reasoning steps containing such tokens are heavily populated with exploratory branches.
Then we construct experiments (experiment group) by retaining the trajectories with the \textit{lowest} proportion of reasoning steps that contain branching keywords (\texttt{bottom K\%}), and corresponding \textit{control experiments} by retaining those with the \textit{highest} proportion (\texttt{top K\%}).
The results are shown in Table \ref{tab:bottom_select}. 
Training on the subsets with fewer branching patterns (experiment groups) consistently enhances reasoning performance over the baseline trained on the full-set of \texttt{DeepSeek-R1} data. 
Notably, using only the \texttt{bottom 50\%} of the data, Qwen3-8B achieves an average performance gain of $2.2\%$, including a remarkable $5.1\%$ absolute improvement on the AIME25 benchmark. 
Similarly, Qwen2.5-7B sees an average improvement of $3.6\%$ with the \texttt{bottom 90\%} part of the data. 
In contrast, the control groups consistently decrease the trained models' performance compared to the original baseline. 
This controlled contrast strongly validates our core insight: the pattern that frequently branching without deep deduction in \texttt{DeepSeek-R1} data acts as structural redundancy during SFT. Filtering out these trajectories prevents the base models from overfitting to inefficient and divergent exploration behaviors, encouraging them to learn more deductive and convergent reasoning paths, which drives better generalization.

\textbf{Ablation Study by Filtering out the Longest Trajectories} 
A natural question can be raised here: "\textit{Do the performance gains simply come from that we filter out those longest reasoning trajectories?}"
In our experiments, we find that the trajectories filtered out with our proposed two proxy metrics have a very limited overlap with the trajectories filtered out with the length metric.
We also carefully conduct an ablation study by filtering out $20\%$ longest CoT trajectories and retraining the base model. 

\begin{table}[t]
\centering
\caption{
Results of the ablation study. We compare the SFT generalization performance of filtering out the longest $20\%$ trajectories with our proposed proxy metric 1 and proxy metric 2, where we filter out the most frequently branching trajectories.
}
\label{tab:ablation_study_length_filter}
%\vspace{-2mm}
\resizebox{\textwidth}{!}{%
\begin{tabular}{@{}l|c|ccccc|c@{}}
\toprule
\textbf{Base Model} & \textbf{Data Source} & \textbf{MATH500} & \textbf{AIME24} & \textbf{AIME25} & \textbf{BeyondAIME} & \textbf{HMMT25} & \textbf{Avg} \\
\midrule
\multirow{4}{*}{\shortstack{Qwen3-8B \\ {\footnotesize (context len: 32K)}}} 
 & full \texttt{DeepSeek-R1} set (original SFT) & $96.8\%$\scriptsize$\pm0.02\%$ & $63.0\%$\scriptsize$\pm0.03\%$ & $54.0\%$\scriptsize$\pm0.03\%$ & $24.5\%$\scriptsize$\pm0.03\%$ & $48.5\%$\scriptsize$\pm0.03\%$ & $57.4\%$ \\
\cmidrule{2-8}
 & filtering out $20\%$ with proxy 1 & \cellcolor{green!25}$97.2\%$\scriptsize$\pm0.02\%$& \cellcolor{green!25}$64.5\%$\scriptsize$\pm0.03\%$& \cellcolor{green!25}$54.7\%$\scriptsize$\pm0.03\%$& \cellcolor{green!25}$25.7\%$\scriptsize$\pm0.03\%$ & \cellcolor{green!25}$51.5\%$\scriptsize$\pm0.03\%$& 
 $\mathbf{58.4\%}$ {\footnotesize \textcolor{green!60!black}{(\textbf{+1.0\%})}} \\
 & filtering out $20\%$ with proxy 2 & \cellcolor{green!25}$97.0\%$\scriptsize$\pm0.02\%$ & \cellcolor{green!25}$64.7\%$\scriptsize$\pm0.03\%$ & \cellcolor{green!25}$55.7\%$\scriptsize$\pm0.03\%$ & \cellcolor{green!25}$27.1\%$\scriptsize$\pm0.03\%$ & \cellcolor{green!25}$52.3\%$\scriptsize$\pm0.03\%$ & $\mathbf{59.4\%}$ {\footnotesize \textcolor{green!60!black}{(\textbf{+2.0\%})}} \\
 \cmidrule{2-8}
   & \textbf{ablation:} filtering out the longest $20\%$ & \cellcolor{red!25}$96.0\%$\scriptsize$\pm0.02\%$ & \cellcolor{red!25}$61.0\%$\scriptsize$\pm0.03\%$ & \cellcolor{red!25}$46.9\%$\scriptsize$\pm0.03\%$ & \cellcolor{red!25}$23.3\%$\scriptsize$\pm0.03\%$ & \cellcolor{red!25}$45.1\%$\scriptsize$\pm0.03\%$ & $54.5\%$ {\footnotesize \textcolor{red!60!black}{(\textbf{-2.9\%})}} \\
\midrule
\midrule
\multirow{4}{*}{\shortstack{Qwen2.5-7B \\ {\footnotesize (context len: 32K)}}} 
 & full \texttt{DeepSeek-R1} set (original SFT) & $95.0\%$\scriptsize$\pm0.02\%$ & $52.1\%$\scriptsize$\pm0.03\%$ & $44.8\%$\scriptsize$\pm0.03\%$ & $16.7\%$\scriptsize$\pm0.02\%$ & $38.7\%$\scriptsize$\pm0.03\%$ & $49.5\%$ \\
   \cmidrule{2-8}
 & filtering out $20\%$ with proxy 1 & \cellcolor{green!25}$95.8\%$\scriptsize$\pm0.02\%$& \cellcolor{green!25}$52.2\%$\scriptsize$\pm0.03\%$& \cellcolor{red!25}$44.4\%$\scriptsize$\pm0.03\%$& \cellcolor{red!25}$16.2\%$\scriptsize$\pm0.02\%$ & \cellcolor{green!25}$40.0\%$\scriptsize$\pm0.03\%$& 
 $\mathbf{49.7\%}$ {\footnotesize \textcolor{green!60!black}{(\textbf{+0.2\%})}} \\
 & filtering out $20\%$ with proxy 2 & \cellcolor{red!25}$94.6\%$\scriptsize$\pm0.02\%$ & \cellcolor{green!25}$55.5\%$\scriptsize$\pm0.03\%$ & \cellcolor{green!25}$47.2\%$\scriptsize$\pm0.03\%$ & \cellcolor{green!25}$18.5\%$\scriptsize$\pm0.02\%$ & \cellcolor{green!25}$40.1\%$\scriptsize$\pm0.03\%$ & $\mathbf{51.2\%}$ {\footnotesize \textcolor{green!60!black}{(\textbf{+1.7\%})}} \\
 \cmidrule{2-8}
  & \textbf{ablation:} filtering out the longest $20\%$ & \cellcolor{red!25}$93.6\%$\scriptsize$\pm0.02\%$ & \cellcolor{red!25}$44.7\%$\scriptsize$\pm0.03\%$ & \cellcolor{red!25}$38.0\%$\scriptsize$\pm0.03\%$ & \cellcolor{red!25}$13.1\%$\scriptsize$\pm0.02\%$ & \cellcolor{red!25}$32.7\%$\scriptsize$\pm0.03\%$ & $44.4\%$ {\footnotesize \textcolor{red!60!black}{(\textbf{-5.1\%})}} \\
 \midrule
 \midrule
\multirow{4}{*}{\shortstack{Qwen2.5-7B \\ {\footnotesize (context len: 96K)}}} 
 & full \texttt{DeepSeek-R1} set (original SFT)  & $93.6\%$\scriptsize$\pm0.02\%$ & $61.7\%$\scriptsize$\pm0.03\%$ & $53.4\%$\scriptsize$\pm0.03\%$ & $27.5\%$\scriptsize$\pm0.02\%$ & $38.3\%$\scriptsize$\pm0.03\%$ & $54.9\%$ \\
\cmidrule{2-8}
 & filtering out $20\%$ with proxy 1 & \cellcolor{green!25}$95.6\%$\scriptsize$\pm0.02\%$& \cellcolor{green!25}$64.8\%$\scriptsize$\pm0.03\%$& \cellcolor{red!25}$53.1\%$\scriptsize$\pm0.03\%$& \cellcolor{green!25}$29.2\%$\scriptsize$\pm0.03\%$ & \cellcolor{green!25}$39.3\%$\scriptsize$\pm0.03\%$& 
 $\mathbf{56.4\%}$ {\footnotesize \textcolor{green!60!black}{(\textbf{+1.5\%})}} \\
 & filtering out $20\%$ with proxy 2 & \cellcolor{green!25}$95.8\%$\scriptsize$\pm0.02\%$ & \cellcolor{green!25}$65.0\%$\scriptsize$\pm0.03\%$ & \cellcolor{green!25}$56.8\%$\scriptsize$\pm0.03\%$ & \cellcolor{green!25}$32.1\%$\scriptsize$\pm0.03\%$ & \cellcolor{green!25}$40.5\%$\scriptsize$\pm0.03\%$ & $\mathbf{58.0\%}$ {\footnotesize \textcolor{green!60!black}{(\textbf{+3.1\%})}} \\
 \cmidrule{2-8}
  & \textbf{ablation:} filtering out the longest $20\%$ & \cellcolor{green!25}$94.0\%$\scriptsize$\pm0.02\%$ & \cellcolor{red!25}$51.5\%$\scriptsize$\pm0.03\%$ & \cellcolor{red!25}$41.9\%$\scriptsize$\pm0.03\%$ & \cellcolor{red!25}$20.8\%$\scriptsize$\pm0.03\%$ & \cellcolor{red!25}$33.8\%$\scriptsize$\pm0.03\%$ & $48.4\%$ {\footnotesize \textcolor{red!60!black}{(\textbf{-6.5\%})}} \\
\bottomrule
\end{tabular}
}
\vspace{-4mm}
\end{table}

The results of the experiment study are shown in Table~\ref{tab:ablation_study_length_filter}, which indicates that simply filtering out the longest trajectories cannot achieve similar performance gains with our proposed approaches.
This further demonstrates that our main results should not be attributed to filtering out the longest trajectories, but attributed to filtered out trajectories with the most branching reasoning behaviors.
\begin{insightbox}
\textbf{Takeaway:}
filtering out the most frequently branching trajectories from the original \texttt{DeepSeek-R1} data can starkly and consistently improve SFT generalization performance.
\end{insightbox}
\section{Conclusion}
We investigated the generalization discrepancy in Long CoT SFT and demonstrated that the intrinsic reasoning patterns of the training data serve as a critical factor shaping model performance. 
% We first utilize reasoning trajectories from \texttt{DeepSeek-R1} and \texttt{gpt-oss-120b} to fine-tune base models. 
Our multi-faceted analysis revealed that the pattern of frequent, divergent branching without deep deduction in long CoT trajectories introduces structural redundancy, causing student models to be trapped in redundant exploratory branches that hinder them from reaching the correct solutions. 
By removing the most frequently branching trajectories, we significantly and consistently improved the generalization of SFT across benchmarks.
\clearpage
\bibliography{colm2026_conference}
\bibliographystyle{colm2026_conference}
\clearpage
\appendix
\section{Disclosure of LLM Use}
In compliance with the formatting and ethical guidelines of the conference, we disclose the use of Large Language Models (LLMs) in the preparation of this manuscript. Specifically, LLMs were employed strictly as assistive tools for the following tasks: (1) polishing the writing, correcting typos and grammar errors, and resolving \LaTeX\ syntax issues; and (2) assisting in the refinement of \texttt{matplotlib} Python scripts for visualizing data and making figures.

We explicitly state that all core intellectual contributions are entirely the work of the human authors, including the conceptualization of the research ideas, the formulation of the experimental pipeline, the execution of the experiments, and the derivation of the scientific conclusions. \textbf{No LLMs were utilized to generate novel scientific insights or to design the experimental methodologies presented in this paper.}

\section{Limitations}
Our empirical study primarily focuses on complex mathematical reasoning benchmarks, which serve as the standard testbed for current Large Reasoning Models. Future work should investigate whether the observed generalization discrepancy and the efficacy of our structural filtering strategy seamlessly extend to other domains requiring reasoning, such as code generation or agentic tasks. In our work, we adopt reasoning trajectories of two widely-used open-source models, \texttt{DeepSeek-R1-0528} and \texttt{gpt-oss-120b}. The key motivation of this work is to study the reasoning patterns of different LRMs and their effect on the generalization performance of SFT. We leave the discussion of more teacher models~\citep{kimiteam2026kimik25visualagentic,deepseekai2025deepseekv32pushingfrontieropen} in the future work. 
\section{Experimental Setup}
\label{appendix:experiment_setup}
\subsection{SFT Training Setup}
Following~\citet{tian2025not}, we collect a high-quality dataset comprising approximately 500,000 challenging mathematical problems from publicly available datasets including OpenR1-Math-220k\footnote{\url{https://huggingface.co/datasets/open-r1/OpenR1-Math-220k}}, Big-Math-RL-Verified~\citep{albalak2025bigmathlargescalehighqualitymath}, NuminaMath~\citep{numina_math_datasets} and so on. For each problem, we query both \texttt{DeepSeek-R1-0528} and \texttt{gpt-oss-120b} to generate their respective Long CoT trajectories. To rigorously control data quality, we apply a rule-based verification pipeline to ensure that \textit{all trajectories used in our experiments successfully arrive at the correct final answer}. This step guarantees that any observed performance differences stem from the intrinsic structural properties of the reasoning paths rather than factual correctness.
The length (token number) of the reasoning trajectories used for SFT training is strictly controlled under 32k. 
We select four representative (covering different model families including Qwen2.5, Qwen3 and LLaMA3.1, and different model scales from 7B to 32B) open-weight base models as student models: Qwen2.5-7B\footnote{\url{https://huggingface.co/Qwen/Qwen2.5-7B}}~\citep{qwen2025qwen25technicalreport}, Qwen2.5-32B\footnote{\url{https://huggingface.co/Qwen/Qwen2.5-32B}}~\citep{qwen2025qwen25technicalreport}, Llama3.1-8B\footnote{\url{https://huggingface.co/meta-llama/Llama-3.1-8B}}~\citep{grattafiori2024llama3herdmodels}, and Qwen3-8B\footnote{\url{https://huggingface.co/Qwen/Qwen3-8B-Base}}~\citep{yang2025qwen3technicalreport}. Each model undergoes SFT on the \texttt{DeepSeek-R1} and \texttt{gpt-oss-120b} datasets independently. We maintain identical hyperparameter settings and optimize for the approximately same number of training steps across each pair of comparisons. 
For all of the SFT experiments, we use the Adam optimizer~\citep{kingma2014adam} with the $\beta_1$=0.9, $\beta_2$=0.95, gradient clip = 1.0, training sequence length = 32768, the Cosine learning rate scheduler with a warmup fraction = 0.05 and a minimum learning rate = 5e-6. 
For other main training hyperparameters, including train step, (initial) learning rate, global batch size and micro batch size, please refer to Table~\ref{tab:training_hyper_parameters}.
\begin{table}[h]
\centering
\caption{Training hyperparameters for the long CoT SFT experiments in the main table.}
\label{tab:training_hyper_parameters}
\resizebox{\textwidth}{!}{%
\begin{tabular}{@{}c|c|cccc}
\toprule
\textbf{Base Model} & \textbf{Data Source} & \textbf{train step} & \textbf{learning rate} & \textbf{global batch size} & \textbf{micro batch size} \\
\midrule
\multirow{2}{*}{Qwen2.5-7B} 
 & DeepSeek-R1   & $1836$ & $1e-4$ & $128$ & $1$ \\
 & gpt-oss-120b  & $1750$ & $1e-4$ & $128$ & $1$ \\
\midrule
\multirow{2}{*}{Qwen2.5-32B} 
 & DeepSeek-R1   & $3672$ & $8e-5$ & $64$ & $1$ \\
 & gpt-oss-120b  & $3501$ & $8e-5$ & $64$ & $1$ \\
\midrule
\multirow{2}{*}{Llama3.1-8B} 
 & DeepSeek-R1   & $1776$ & $1e-4$ & $128$ & $1$ \\
 & gpt-oss-120b  & $1733$ & $1e-4$ & $128$ & $1$ \\
\midrule
\multirow{2}{*}{Qwen3-8B} 
 & DeepSeek-R1   & $1835$ & $1e-4$ & $128$ & $1$ \\
 & gpt-oss-120b  & $1749$ & $1e-4$ & $128$ & $1$ \\
\bottomrule
\end{tabular}%
}
\end{table}

\subsection{Inference Setup}
The trained models are evaluated on five representative mathematical reasoning benchmarks: MATH500~\citep{hendrycks2021measuring}, AIME24~\citep{aime24}, AIME25~\citep{aime25}, BeyondAIME~\citep{bytedance_seed_2025_beyondaime}, and HMMT25~\citep{balunovic_srimatharena_2025}.
During all inference experiments, we set the temperature = 0.6, the repetition penalty = 1.0, top-k = 20, top-p=0.95.
To ensure consistency with the training phase and accommodate complex reasoning chains, we set the maximum output token limit for each path to 32K tokens by default. We extract the final answer from the model’s output by parsing the content within the last $\backslash$boxed\{\} command. The extracted answer is then compared against the ground truth using the math-verify library\footnote{\url{https://github.com/huggingface/Math-Verify}} to determine correctness.

\subsection{Reasoning Behavior Analysis Setup}
\label{appendix:experiment_setup_reason_behavior}
The detailed definition of four common reasoning behaviors (including \texttt{Propose}, \texttt{Deduce}, \texttt{Verify}, and \texttt{Backtrack}) is provided in Table~\ref{tab:reasoning_behavior_intro}.
The prompt template that is used for annotating reasoning steps is provided in Figure~\ref{figure:annotation_template}. For each annotation, we provide \texttt{DeepSeek-V3.2}~\citep{deepseekai2025deepseekv32pushingfrontieropen} (chat version) with two consecutive reasoning steps (i.e., the ‘previous reasoning step’ and the ‘current reasoning step’) and ask it to judge which behavior the logical transition (from the ‘previous reasoning step’ to the ‘current reasoning step’) most likely belongs to.
We set the temperature = 0. to get the classification result with the largest probability.
In our experiments, we randomly sample $100$ trajectories from the whole training set (for both of \texttt{DeepSeek-R1-0528} data and \texttt{gpt-oss-120b} data, their sampled trajectories corresponds to the same set of problems), resulting in 25978 reasoning steps for \texttt{DeepSeek-R1-0528} and 11316 reasoning steps for \texttt{gpt-oss-120b}, for our analysis. For the experiments with the AIME24 problems, we use the whole set of AIME problems for our analysis.

Besides, we split the reasoning trajectories into reasoning steps following a four-level split procedure: we use the four signal strings ('$\backslash n \backslash n$', '$\backslash n$', '?', '.') one-by-one to split the original trajectories into fine-grained steps.

Note that the design of our prompt template for behavior annotation is not purely out of intuition: we construct the initial version of our prompt based on the prompt templates that are used in~\citet{chen2026molecular},~\citet{gandhi2025cognitive}, and~\citet{jiang2025makes}, then we leverage a human-in-the-loop framework to continuously refine our prompt template. 
For instance, we find that even the state-of-the-art LLM annotators~\citep{deepseekai2025deepseekv32pushingfrontieropen,kimiteam2026kimik25visualagentic} sometimes mistakenly use the label of the "previous step" to annotate the "current step". Hence we add a critical rule for annotation to guide LLM annotators to concentrate on the action in the "current step".
To rigorously validate the reliability of our LLM-based annotation pipeline, we conducted a human evaluation on a random subset of 200 reasoning steps. The LLM annotations achieved a high agreement with expert human labels (accuracy = 92.5\%), confirming the semantic stability of our behavioral taxonomy.
\begin{table}[htbp]
    \centering
    \caption{Definition of four common reasoning behaviors in long CoT reasoning trajectories.}
    \label{tab:reasoning_behavior_intro}
    
    \begin{tabularx}{\textwidth}{>{\centering\arraybackslash}m{2cm}| >{\centering\arraybackslash}X}
        \toprule
        \textbf{Behavior} 
            & \textbf{Definition} \\
        \midrule
        \texttt{Propose}
            & Explore a new idea, set up a hypothesis, or suggest an alternative path. It represents the ``divergent'' phase of reasoning. \\
        \midrule
        \texttt{Deduce}
            & Execute a mathematical operation or make a direct logical inference based on the \textit{immediately preceding} established facts or the current hypothesis. It represents the ``convergent, linear'' phase. \\
        \midrule
        \texttt{Verify}
            & Pause the forward progression to double-check an intermediate calculation, verify a condition, or assess if the current path makes sense. \\
        \midrule
        \texttt{Backtrack}
            & Explicitly realize an error or a dead end, reject the current reasoning branch, and retreat to a previous state or prepares to start over. \\
        \bottomrule
    \end{tabularx}
\end{table}

\begin{figure*}[ht]
\centering
\begin{tcolorbox}[width=0.97\textwidth, colback=white,colframe=black,title=Reasoning Behavior Annotation Template]
\scriptsize
You are an expert cognitive scientist and logician analyzing the internal Long "Chain of Thought" (CoT) reasoning trajectories of Large Language Models.

Your task is to classify the given reasoning step into one of four strictly defined cognitive actions: `[Propose, Deduce, Verify, Backtrack]`.

\vspace{1em}

\textbf{User Prompt:}

You are provided with two consecutive reasoning steps, [PREVIOUS STEP] and [CURRENT STEP], in a Long CoT reasoning trace. 

Please analyze the [CURRENT STEP] and assign ONE of the following four labels.

Note that [PREVIOUS STEP] here is providing you with some necessary context to assist you for analyzing.

\vspace{1em}

\textbf{Label Definitions \& Constraints:}

\vspace{1em}

\textbf{1. Propose (Hypothesis \& Exploration)}

- \textbf{Definition:} The model is exploring a new idea, setting up a hypothesis, or suggesting an alternative path. It represents the "divergent" phase of reasoning.

- \textbf{Significance:} High frequency of this label indicates a highly exploratory, tree-like search structure.

\vspace{1em}

\textbf{2. Deduce (Sequential Deduction)}

- \textbf{Definition:} The model is executing a mathematical operation or making a direct logical inference based on the *immediately preceding* established facts or the current hypothesis. It represents the "convergent, linear" phase.

- \textbf{Significance:} Continuous sequences of this label indicate a dense, high-dependency deductive chain.

\vspace{1em}

\textbf{3. Verify (Self-Reflection \& Checking)}

- \textbf{Definition:} The model pauses its forward progression to double-check an intermediate calculation, verify a condition, or assess if the current path makes sense \textit{without yet abandoning it}.

\vspace{1em}

\textbf{4. Backtrack (Error Correction \& Path Abandonment)}

- \textbf{Definition:} The model explicitly realizes an error or a dead end, rejects the current reasoning branch, and retreats to a previous state or prepares to start over.

\vspace{1em}

\textbf{Critical Rules For Annotation:}

\vspace{1em}

\textbf{1. Focus STRICTLY on the ACTION in [CURRENT STEP].} 

The [PREVIOUS STEP] is strictly for context. Do NOT assign a label based on the tone or action of the [PREVIOUS STEP].

\vspace{1em}

\textbf{2. The "Pivot vs. Progress" Test (Crucial for distinguishing Propose vs. Deduce).}

- \textbf{Progress (Label $\rightarrow$ Deduce):} If [CURRENT STEP] simply executes the math, unpacks the logic, or states the direct consequence of the [PREVIOUS STEP] (e.g., solving the equation just proposed), it is making forward progress. *Key signs: "Thus", "So", "Which means", "That would change...", or direct mathematical formulas.*
  
- \textbf{Pivot (Label $\rightarrow$ Propose):} If [CURRENT STEP] shifts the focus to a NEW angle, introduces a NEW speculation, or brainstorms a different aspect of the problem, it is opening a new branch. *Key signs: The explicit use of words like "Perhaps...", "Alternatively...", "What if...", "Another way..." in the [CURRENT STEP].*

\vspace{1em}

\textbf{Output Format:}

Return exactly one label and nothing else:

'$<$action$>$' (action here can be 'Propose', 'Deduce', 'Verify', 'Backtrack'.)

\vspace{1em}

\textbf{Here is the reasoning step to analyze:}

[PREVIOUS STEP]:

\{\textcolor{blue}{put the previous reasoning step here}\}

[CURRENT STEP]:

\{\textcolor{blue}{put the current reasoning step here}\}
\end{tcolorbox}
\caption{The prompt template for annotating reasoning steps with four behavior labels.}
\label{figure:annotation_template}
\end{figure*}

\section{More Experiment Results}
\label{appendix:more_experiment_results}
\subsection{Additional Token-level SFT Loss Analysis Results}
\begin{figure}[ht]
    \centering
    \captionsetup[subfigure]{justification=centering} 
    % 第一行
    \begin{subfigure}[t]{0.32\textwidth}
        \centering
        \includegraphics[width=\textwidth]{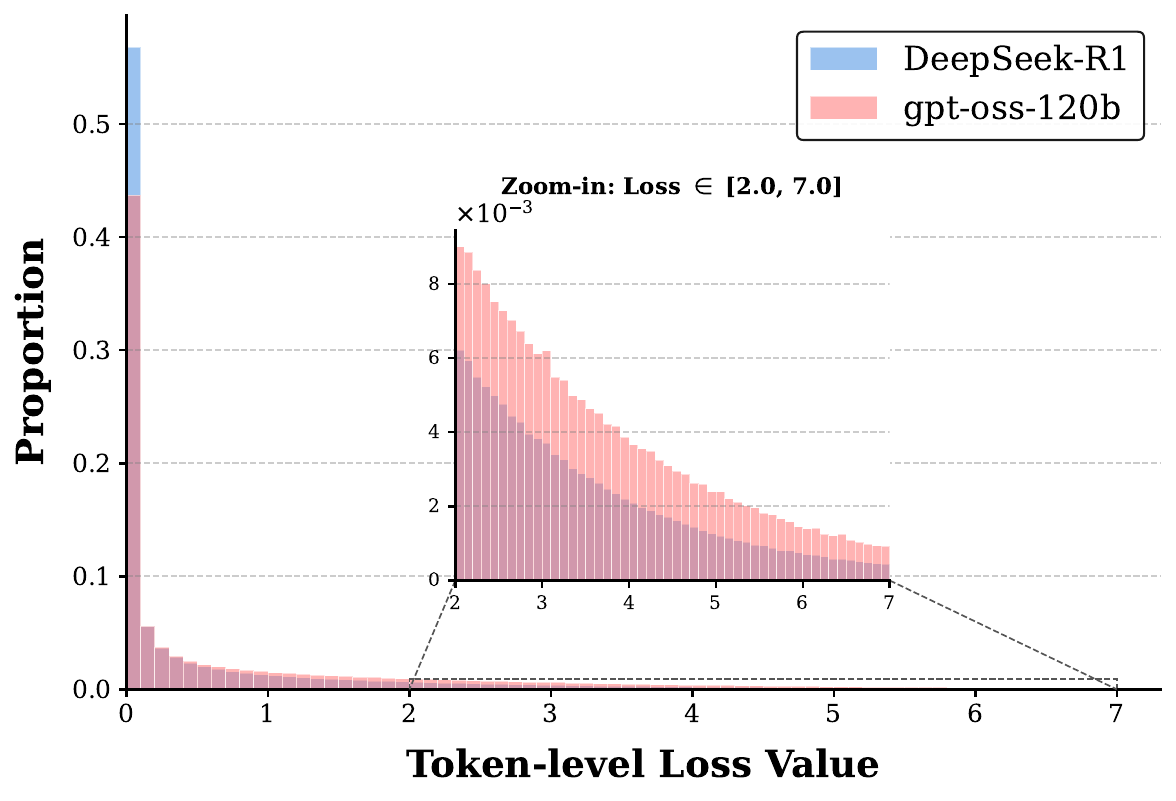}
        \caption{Qwen3-8B’s token-level loss distribution before SFT.}
    \end{subfigure}
    \begin{subfigure}[t]{0.32\textwidth}
        \centering
        \includegraphics[width=\textwidth]{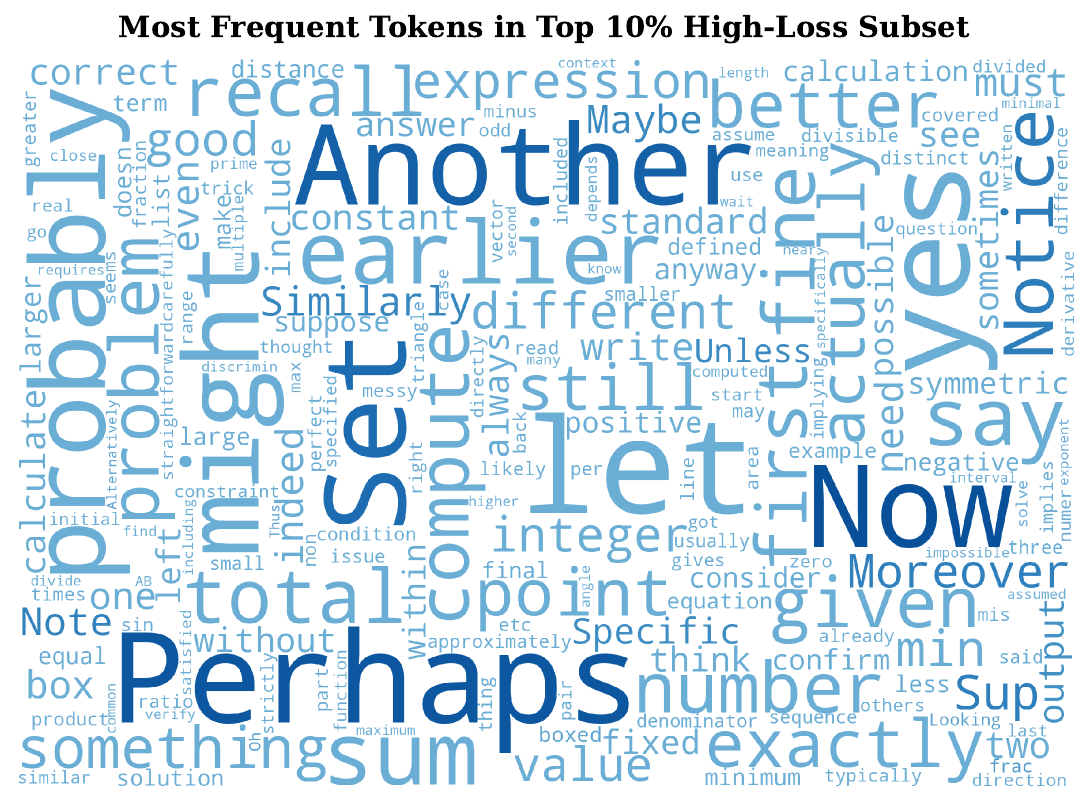}
        \caption{Most frequent tokens in the \texttt{DeepSeek-R1} data before SFT.}
    \end{subfigure}
    \begin{subfigure}[t]{0.32\textwidth}
        \centering
        \includegraphics[width=\textwidth]{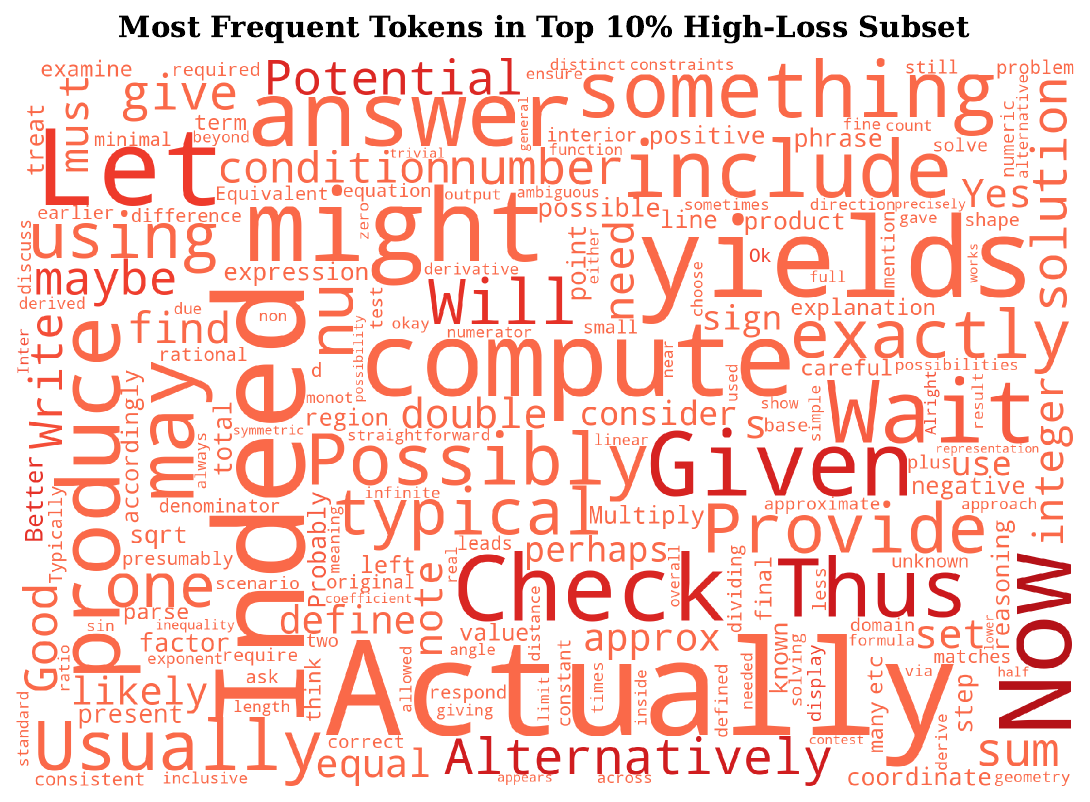}
        \caption{Most frequent tokens in the \texttt{gpt-oss-120b} data before SFT.}
    \end{subfigure}
    \begin{subfigure}[t]{0.32\textwidth}
        \centering
        \includegraphics[width=\textwidth]{figure/loss_distribution_of_two_trained_models.qwen3_8b.pdf}
        \caption{Qwen3-8B’s token-level loss distribution after SFT.}
    \end{subfigure}
    \begin{subfigure}[t]{0.32\textwidth}
        \centering
        \includegraphics[width=\textwidth]{figure/qwen3_8b_dpsk_r1_trained_model_word_cloud.pdf}
        \caption{Most frequent tokens in the \texttt{DeepSeek-R1} data after SFT.}
    \end{subfigure}
    \begin{subfigure}[t]{0.32\textwidth}
        \centering
        \includegraphics[width=\textwidth]{figure/qwen3_8b_oss_120b_trained_model_word_cloud.pdf}
        \caption{Most frequent tokens in the \texttt{gpt-oss-120b} data after SFT.}
    \end{subfigure}
    \caption{Token-level SFT loss analysis for the Qwen3-8B model. (a) and (d) show the token-level loss distribution before and after the SFT training, where \textcolor{blue}{blue}/\textcolor{red}{red} bars represent experiments with \texttt{DeepSeek-R1}/\texttt{gpt-oss-120b}-generated trajectories, respectively. (b, c) and (e, f) are word clouds for the most frequent tokens in the top $10\%$ token-level loss token subset of the (\texttt{DeepSeek-R1}, \texttt{gpt-oss-120b}) data before and after SFT, respectively.}
    \label{fig:token-level-loss-analysis-qwen3-8b}
\vspace{-5mm}
\end{figure}

\begin{figure}[ht]
    \centering
    \captionsetup[subfigure]{justification=centering} 
    % 第一行
    \begin{subfigure}[t]{0.32\textwidth}
        \centering
        \includegraphics[width=\textwidth]{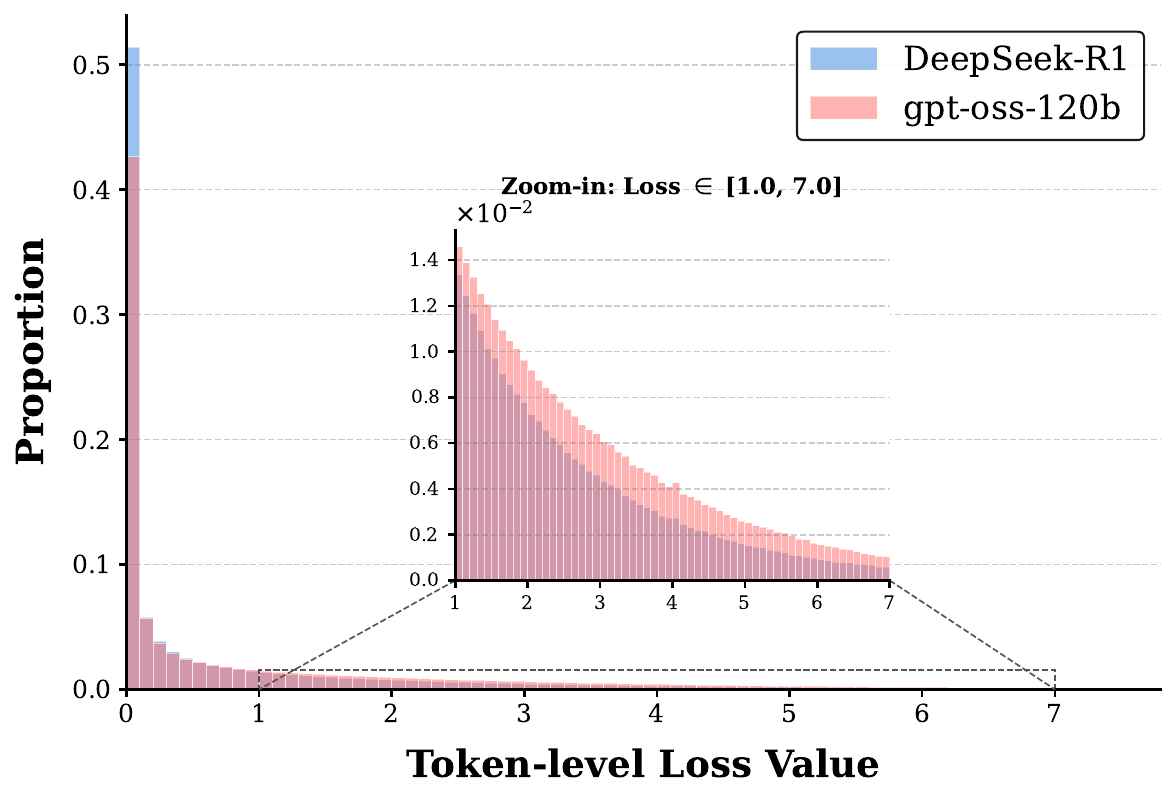}
        \caption{Qwen2.5-7B’s token-level loss distribution before SFT.}
    \end{subfigure}
    \begin{subfigure}[t]{0.32\textwidth}
        \centering
        \includegraphics[width=\textwidth]{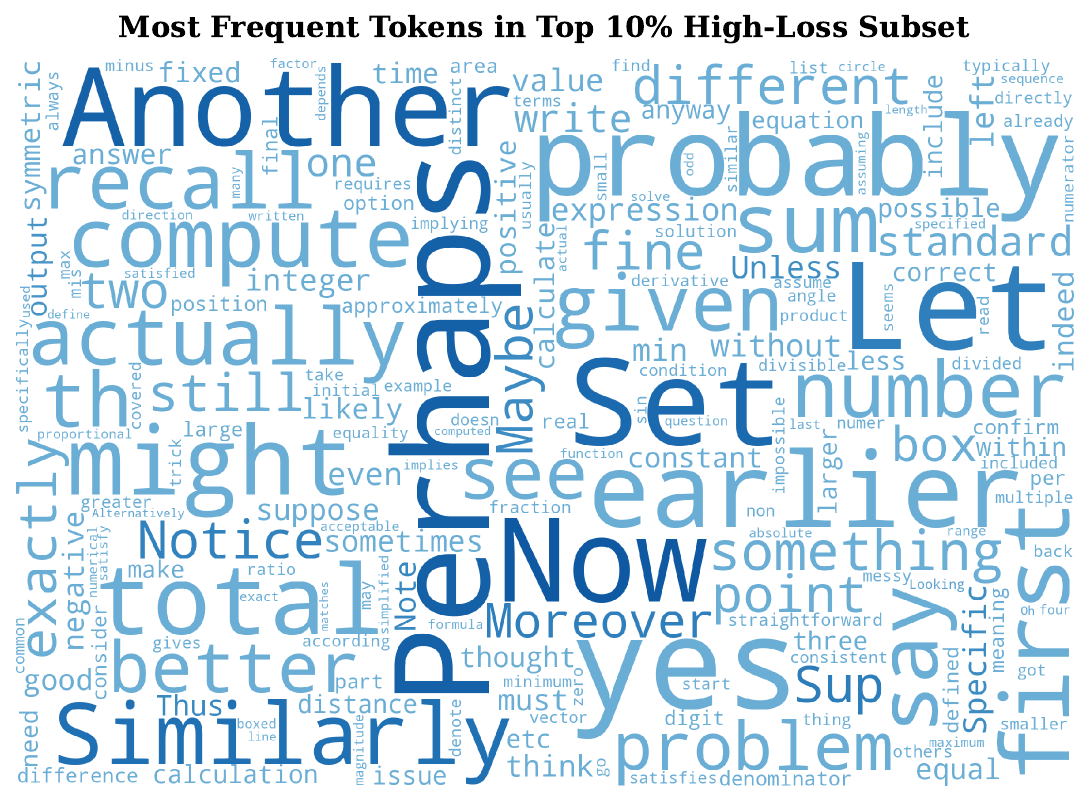}
        \caption{Most frequent tokens in the \texttt{DeepSeek-R1} data before SFT.}
    \end{subfigure}
    \begin{subfigure}[t]{0.32\textwidth}
        \centering
        \includegraphics[width=\textwidth]{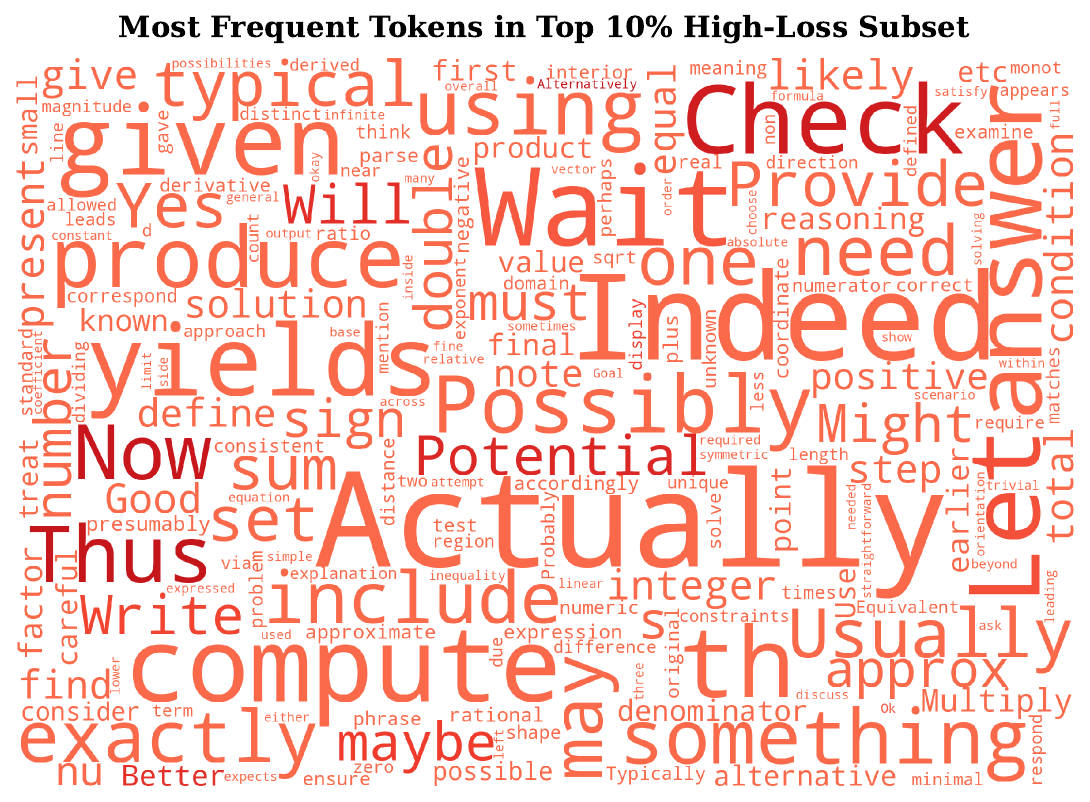}
        \caption{Most frequent tokens in the \texttt{gpt-oss-120b} data before SFT.}
    \end{subfigure}
    \begin{subfigure}[t]{0.32\textwidth}
        \centering
        \includegraphics[width=\textwidth]{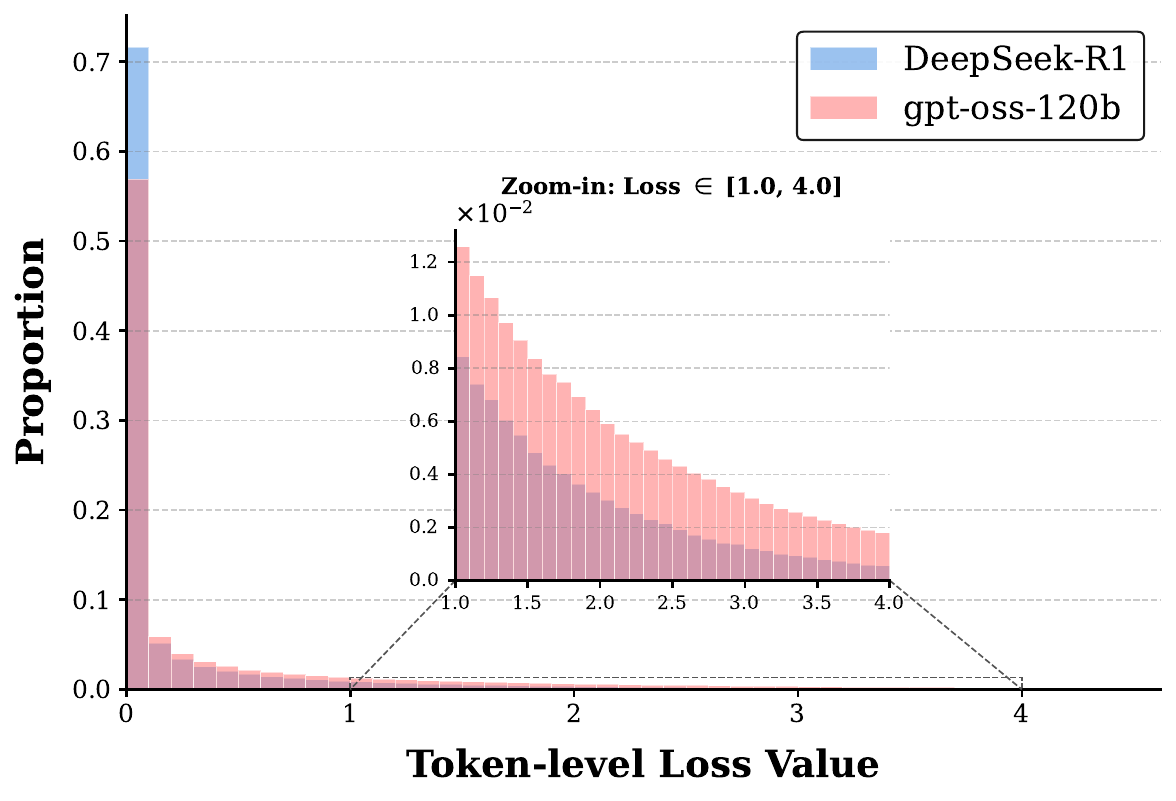}
        \caption{Qwen2.5-7B’s token-level loss distribution after SFT.}
    \end{subfigure}
    \begin{subfigure}[t]{0.32\textwidth}
        \centering
        \includegraphics[width=\textwidth]{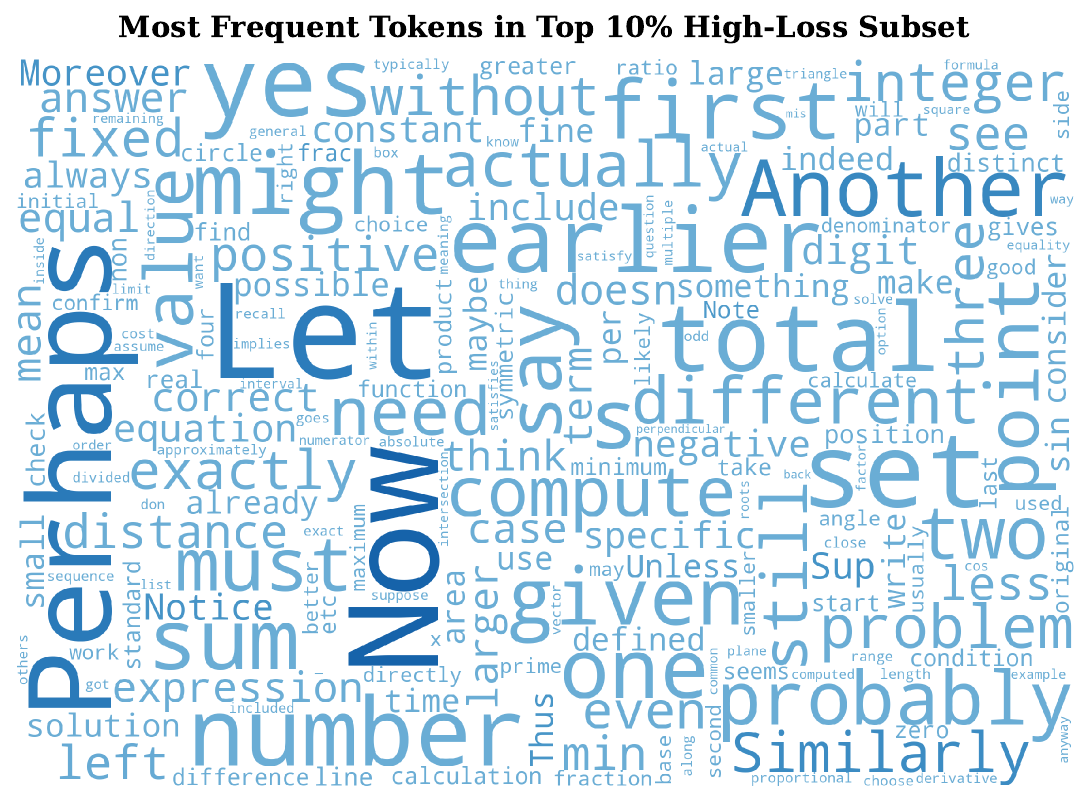}
        \caption{Most frequent tokens in the \texttt{DeepSeek-R1} data after SFT.}
    \end{subfigure}
    \begin{subfigure}[t]{0.32\textwidth}
        \centering
        \includegraphics[width=\textwidth]{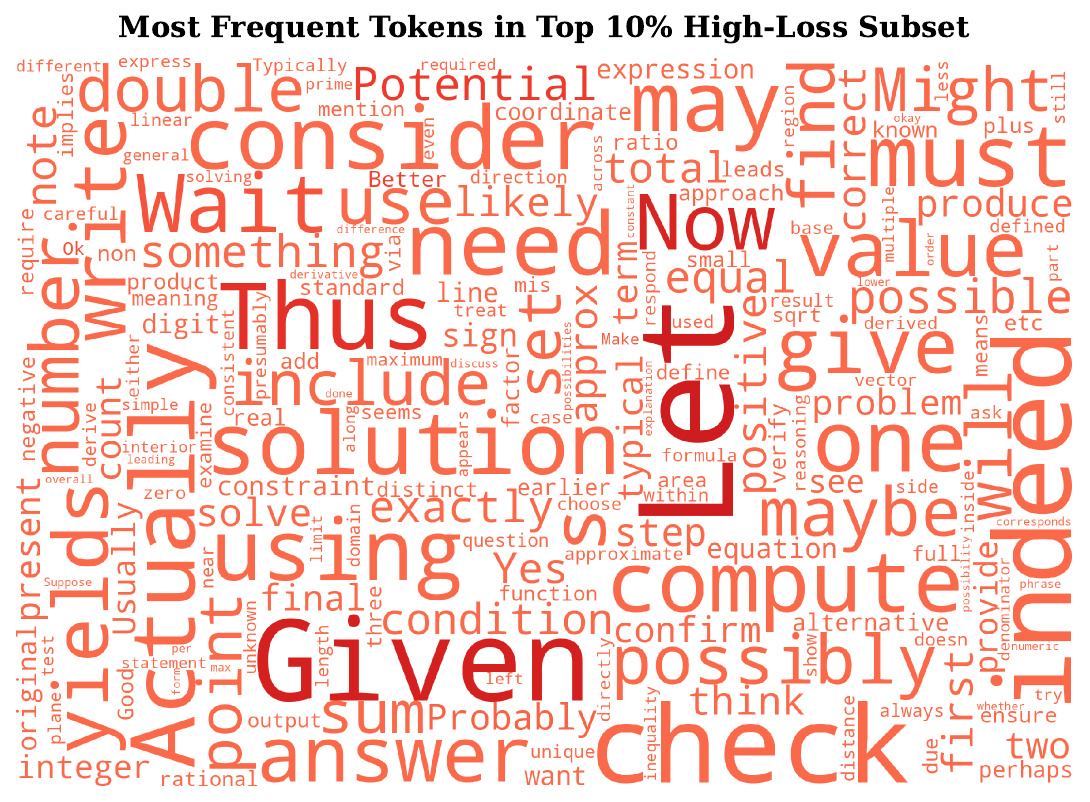}
        \caption{Most frequent tokens in the \texttt{gpt-oss-120b} data after SFT.}
    \end{subfigure}
    \caption{Token-level SFT loss analysis for the Qwen2.5-7B model. (a) and (d) show the token-level loss distribution before and after the SFT training, where \textcolor{blue}{blue} and \textcolor{red}{red} bars correspond to using \texttt{DeepSeek-R1} and \texttt{gpt-oss-120b} reasoning trajectories to train the model and calculate loss, respectively. (b) and (e) are word clouds for the most frequent tokens in the top $10\%$ token-level loss token subset of the \texttt{DeepSeek-R1-0528} generated data before and after the SFT training. (c) and (f) are word clouds for the most frequent tokens in the top $10\%$ token-level loss token subset of the \texttt{gpt-oss-120b} generated data before and after the SFT training.}
    \label{fig:token-level-loss-analysis-qwen25-7b}
\end{figure}

\begin{figure}[ht]
    \centering
    \captionsetup[subfigure]{justification=centering} 
    % 第一行
    \begin{subfigure}[t]{0.32\textwidth}
        \centering
        \includegraphics[width=\textwidth]{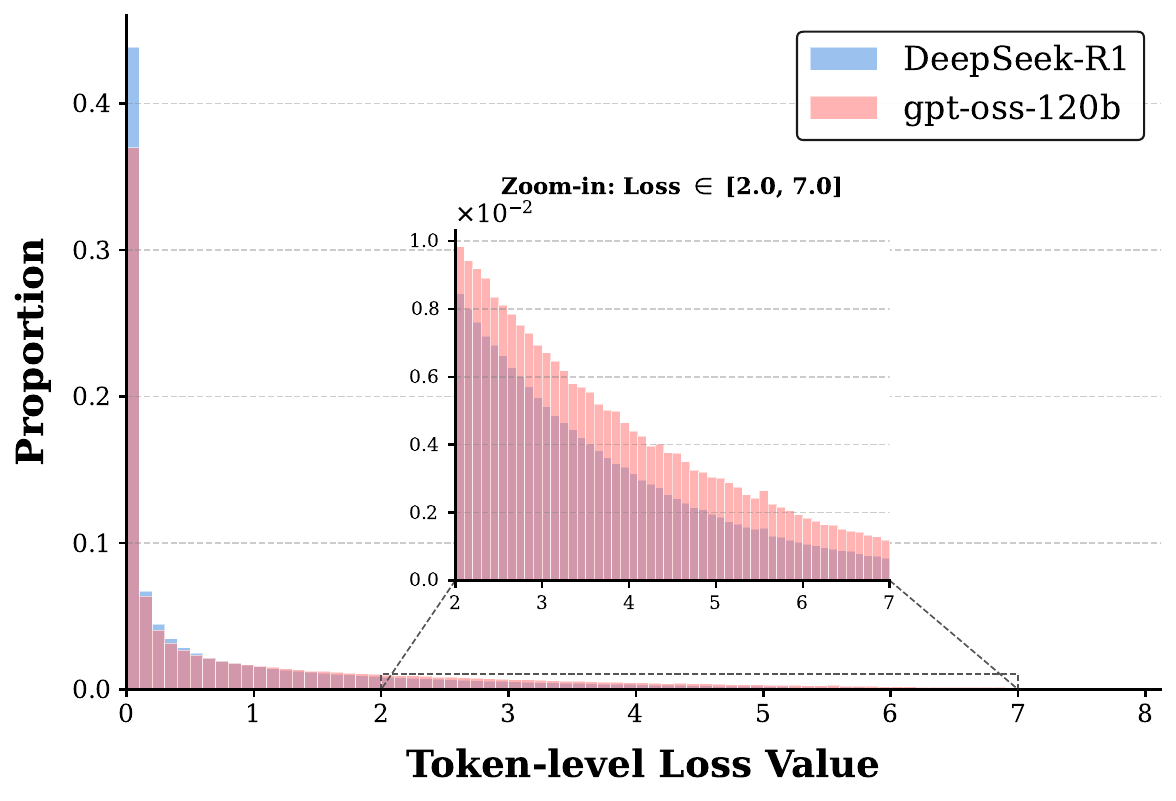}
        \caption{Llama3.1-8B’s token-level loss distribution before SFT.}
    \end{subfigure}
    \begin{subfigure}[t]{0.32\textwidth}
        \centering
        \includegraphics[width=\textwidth]{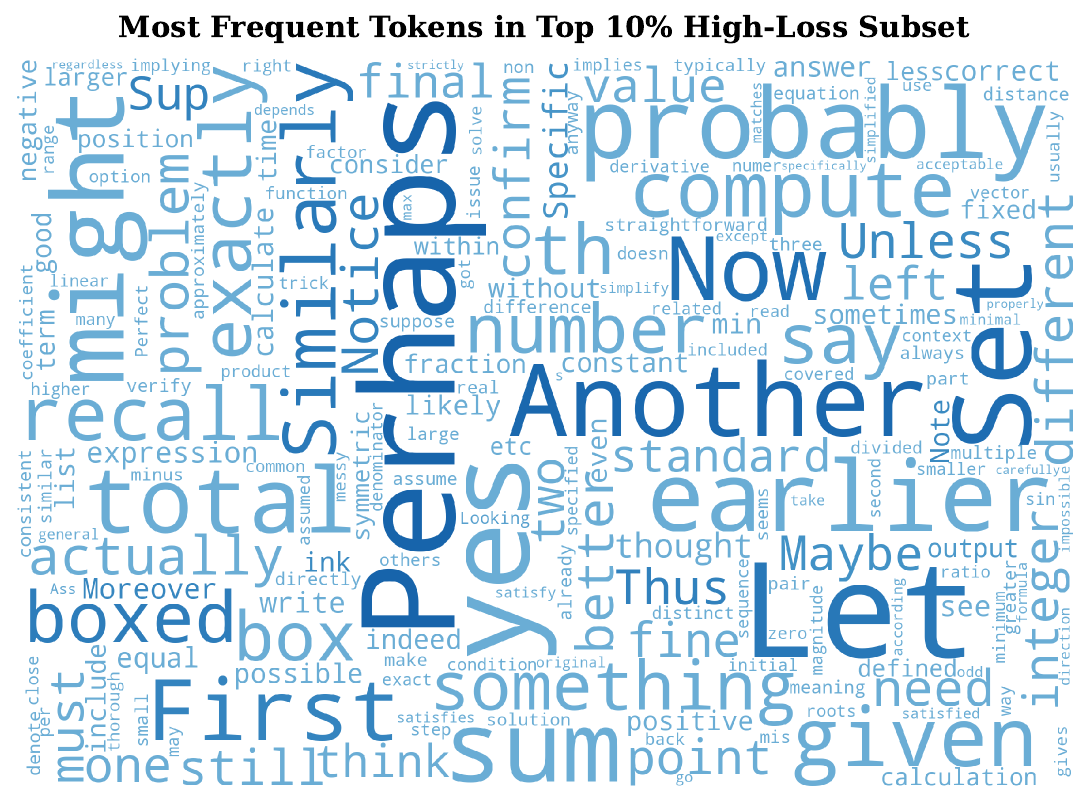}
        \caption{Most frequent tokens in the \texttt{DeepSeek-R1} data before SFT.}
    \end{subfigure}
    \begin{subfigure}[t]{0.32\textwidth}
        \centering
        \includegraphics[width=\textwidth]{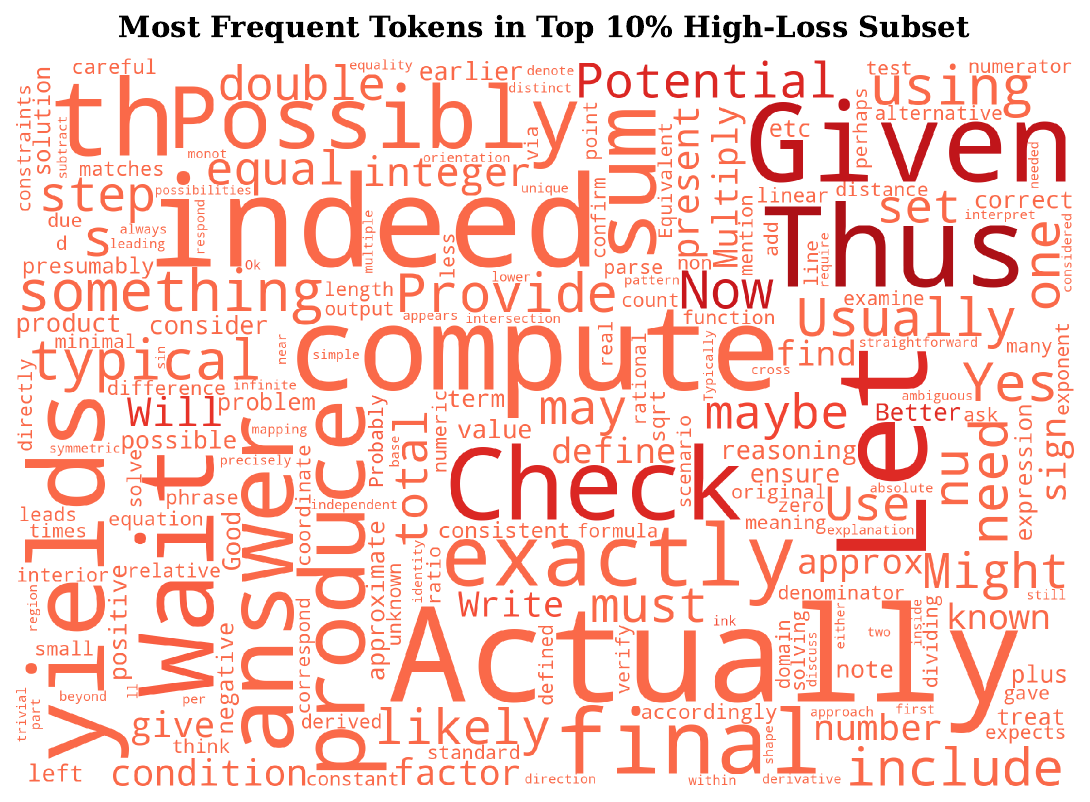}
        \caption{Most frequent tokens in the \texttt{gpt-oss-120b} data before SFT.}
    \end{subfigure}
    \begin{subfigure}[t]{0.32\textwidth}
        \centering
        \includegraphics[width=\textwidth]{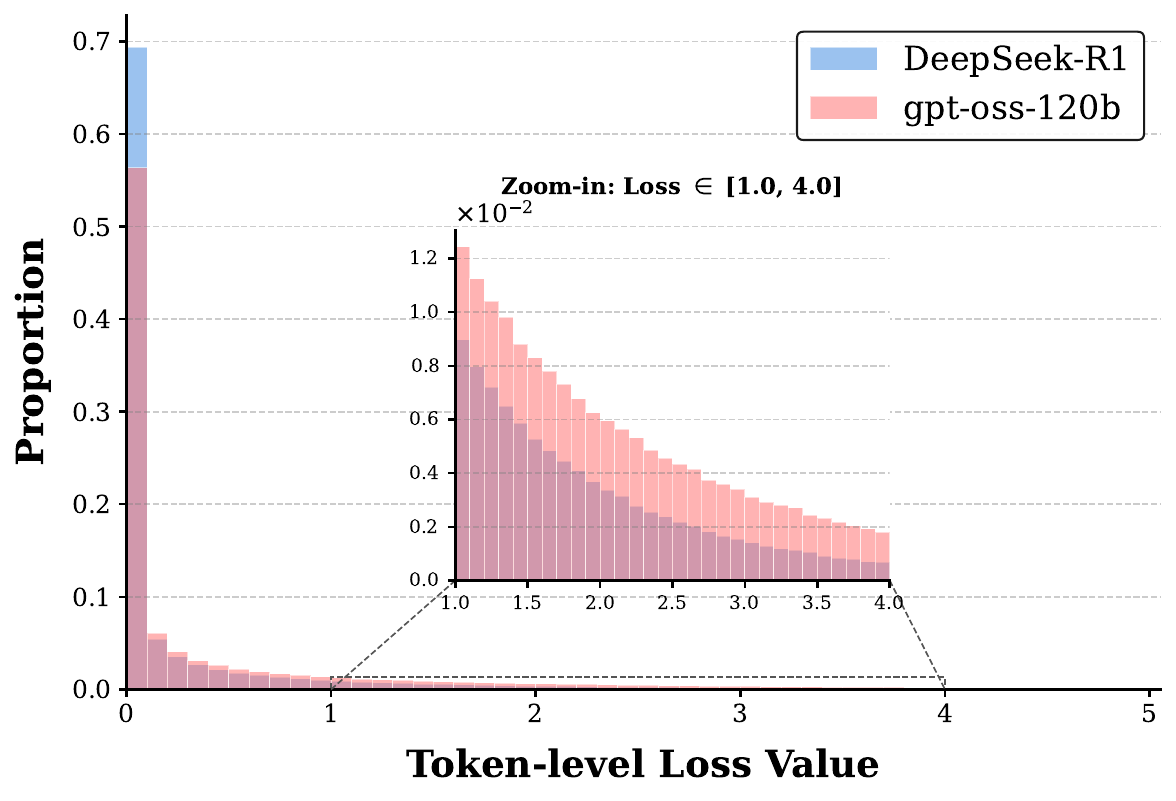}
        \caption{Llama3.1-8B’s token-level loss distribution after SFT.}
    \end{subfigure}
    \begin{subfigure}[t]{0.32\textwidth}
        \centering
        \includegraphics[width=\textwidth]{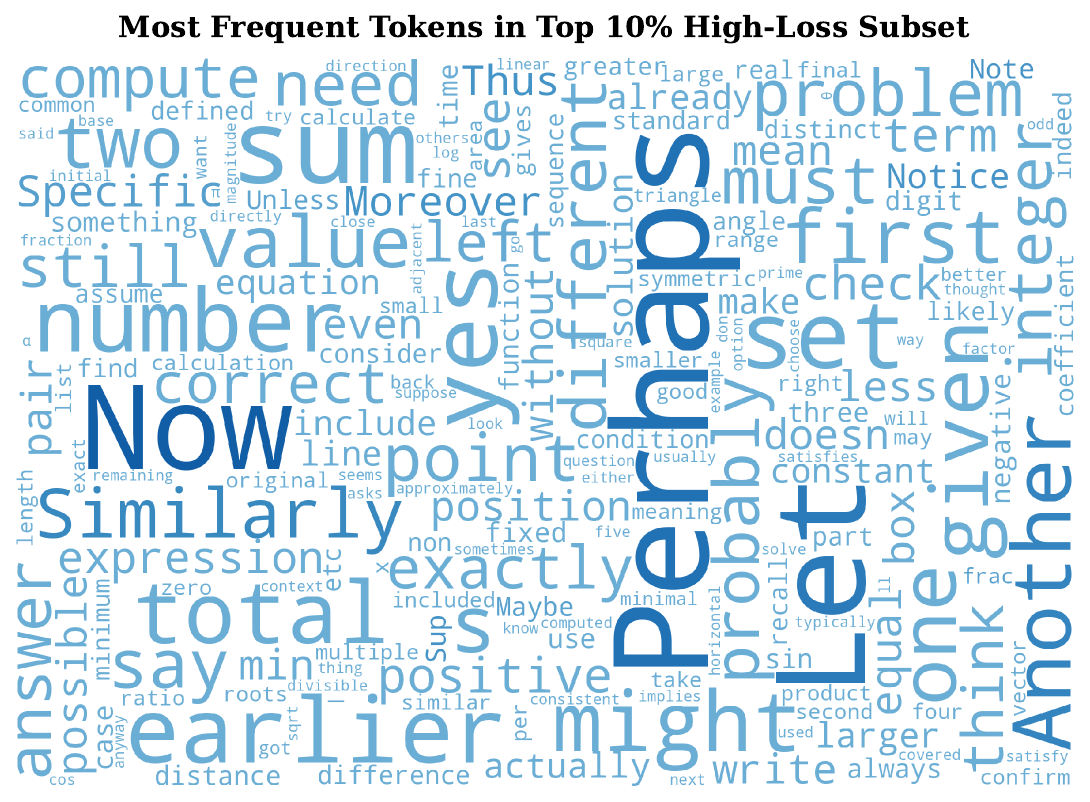}
        \caption{Most frequent tokens in the \texttt{DeepSeek-R1} data after SFT.}
    \end{subfigure}
    \begin{subfigure}[t]{0.32\textwidth}
        \centering
        \includegraphics[width=\textwidth]{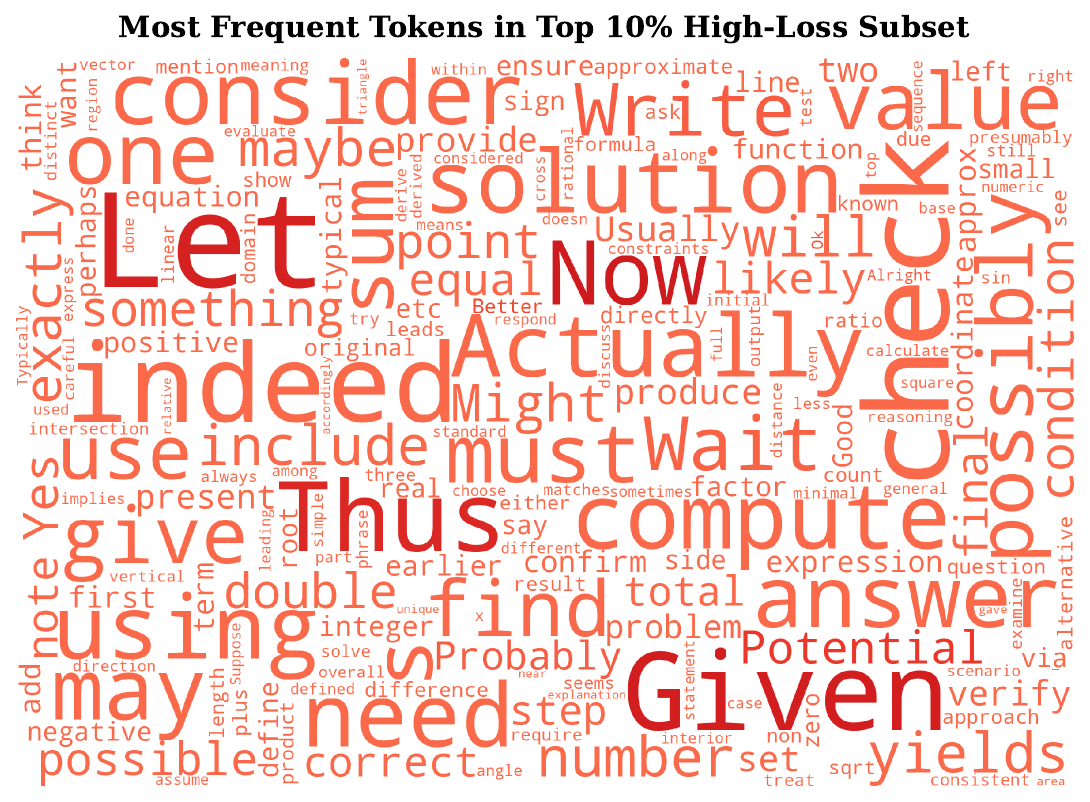}
        \caption{Most frequent tokens in the \texttt{gpt-oss-120b} data after SFT.}
    \end{subfigure}
    \caption{Token-level SFT loss analysis for the Llama3.1-8B model. (a) and (d) show the token-level loss distribution before and after the SFT training, where \textcolor{blue}{blue} and \textcolor{red}{red} bars correspond to using \texttt{DeepSeek-R1} and \texttt{gpt-oss-120b} reasoning trajectories to train the model and calculate loss, respectively. (b) and (e) are word clouds for the most frequent tokens in the top $10\%$ token-level loss token subset of the \texttt{DeepSeek-R1-0528} generated data before and after the SFT training. (c) and (f) are word clouds for the most frequent tokens in the top $10\%$ token-level loss token subset of the \texttt{gpt-oss-120b} generated data before and after the SFT training.}
    \label{fig:token-level-loss-analysis-llama31-8b}
\end{figure}

We show the additional token-level SFT loss analysis results with Qwen3-8B, Qwen2.5-7B, and Llama3.1-8B in Figure~\ref{fig:token-level-loss-analysis-qwen3-8b}, Figure~\ref{fig:token-level-loss-analysis-qwen25-7b}, and Figure~\ref{fig:token-level-loss-analysis-llama31-8b}, respectively.
The statistical results are \textbf{highly aligned with} the results with Qwen3-8B shown in the main context (Figure~\ref{fig:token-level-loss-analysis}).

\subsection{Additional Reasoning Behavior Analysis Results}
\begin{figure}[ht]
    \centering
    \captionsetup[subfigure]{justification=centering} 
    % 第一行
    \begin{subfigure}[t]{0.28\textwidth}
        \centering
        \includegraphics[width=\textwidth]{figure/reasoning_step_distribution.train.pdf}
        \caption{Distribution for training data.}
    \end{subfigure}
    \begin{subfigure}[t]{0.7\textwidth}
        \centering
        \includegraphics[width=\textwidth]{figure/transition_matrix.train.pdf}
        \caption{Transition matrix for reasoning trajectories of training.}
    \end{subfigure}
    \begin{subfigure}[t]{0.28\textwidth}
        \centering
        \includegraphics[width=\textwidth]{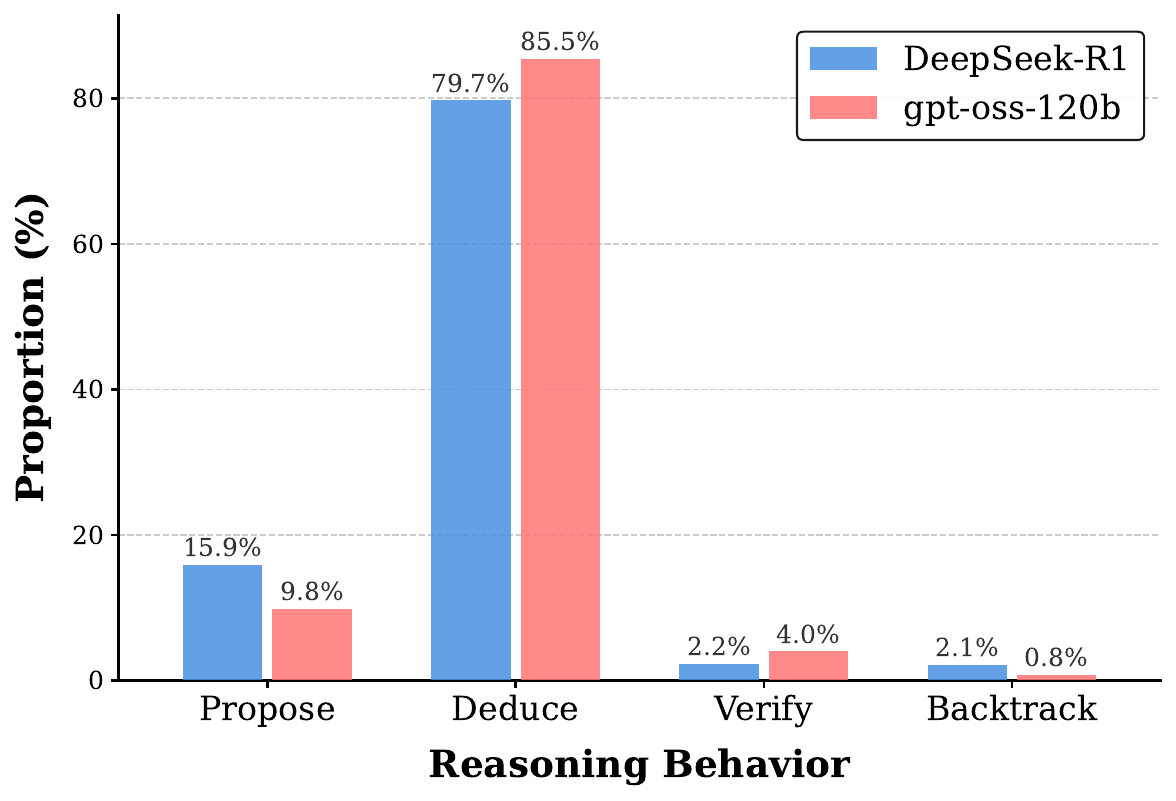}
        \caption{Distribution for AIME24 solutions.}
    \end{subfigure}
    \begin{subfigure}[t]{0.7\textwidth}
        \centering
        \includegraphics[width=\textwidth]{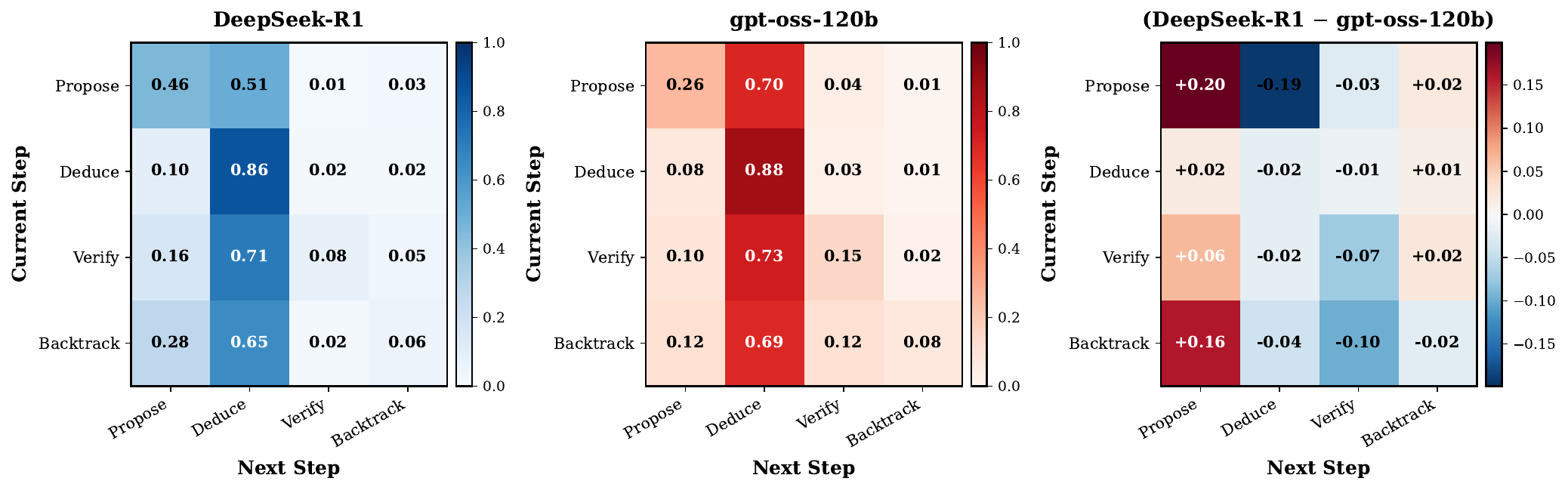}
        \caption{Transition matrix for the generated AIME24 solutions.}
    \end{subfigure}
    \caption{Reasoning behavior analysis. Reasoning behavior distribution ((a) and (c)) and transition matrix ((b) and (d)) for reasoning trajectories collected for the SFT training and generated for AIME testing problems with the trained Llama3.1-8B base model.}
    \label{fig:reason_behavior_distribution_llama31_8b}
\end{figure}

The additional reasoning behavior analysis results with Llama3.1-8B are shown in Figure~\ref{fig:reason_behavior_distribution_llama31_8b}.
The statistical results are \textbf{highly aligned with} the results with Qwen3-8B shown in the main context (Figure~\ref{fig:reason_behavior_distribution_main_text}).

\subsection{Additional Random Reasoning Step Deletion Results}
\begin{figure}[ht]
    \centering
    \captionsetup[subfigure]{justification=centering} 
    % 第一行
    \begin{subfigure}[t]{0.32\textwidth}
        \centering
        \includegraphics[width=\textwidth]{figure/random_del_10p_reason_step_perf_change_bar_qwen3-8b.pdf}
        \caption{Qwen3-8B, delete 10\% steps.}
    \end{subfigure}
    % \hspace{0.04\textwidth}  
    \begin{subfigure}[t]{0.32\textwidth}
        \centering
        \includegraphics[width=\textwidth]{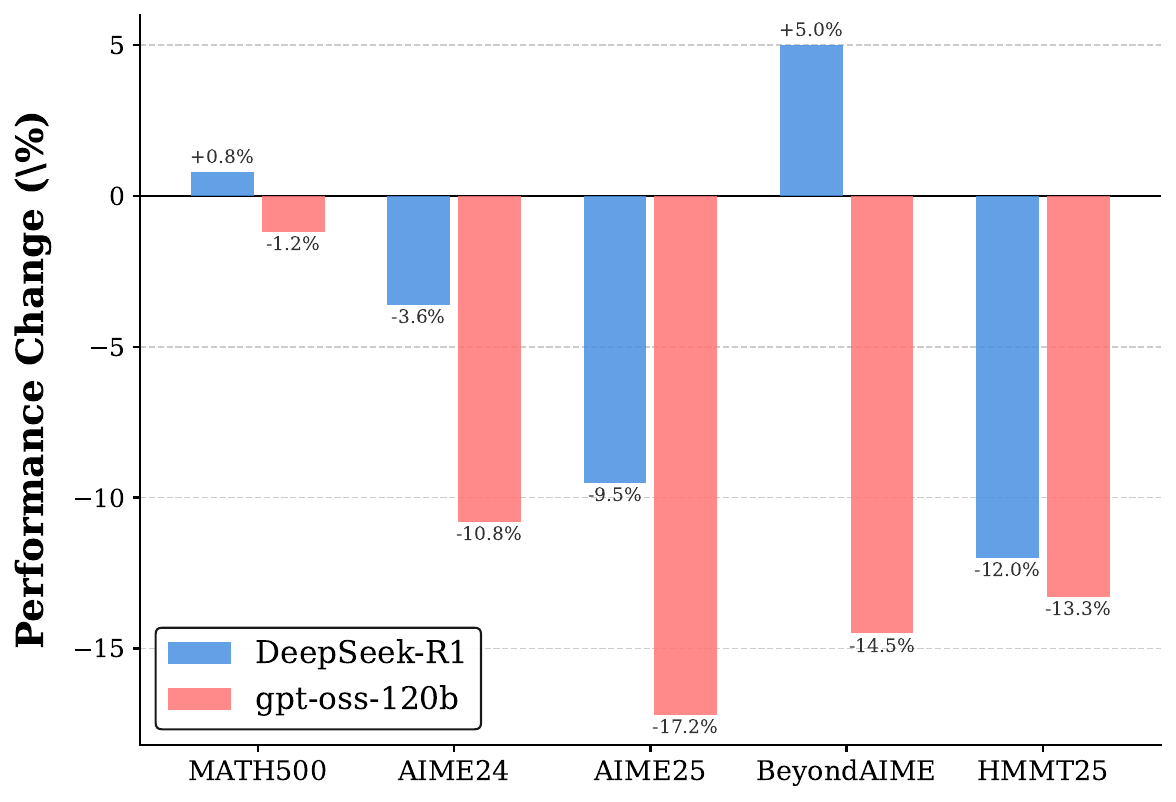}
        \caption{Qwen3-8B, delete 20\% steps.}
    \end{subfigure}
    % \hspace{0.04\textwidth}  
    \begin{subfigure}[t]{0.32\textwidth}
        \centering
        \includegraphics[width=\textwidth]{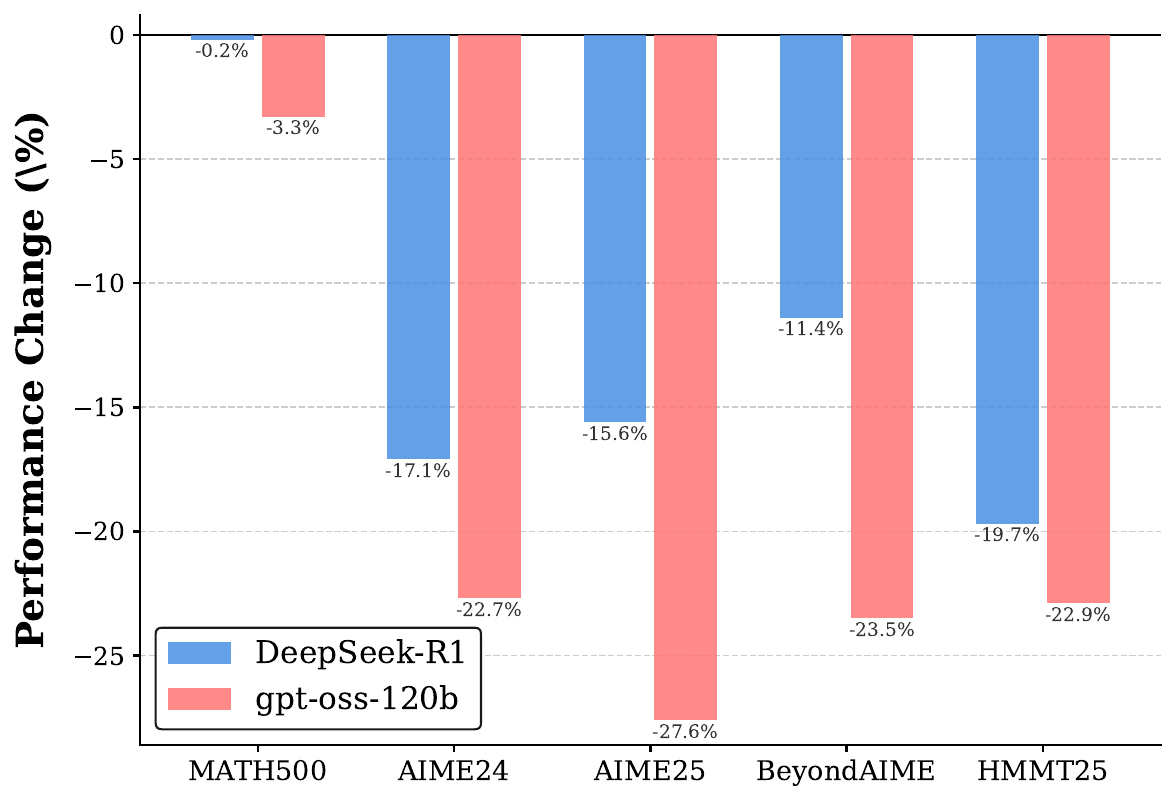}
        \caption{Qwen3-8B, delete 30\% steps.}
    \end{subfigure}
    \caption{Performance change ratio on five benchmarks (MATH500, AIME24/25, BeyondAIME, HMMT25) after randomly deleting 10\%/20\%/30\% reasoning steps in each training datum in the Qwen3-8B SFT experiments. \textcolor{blue}{Blue}/\textcolor{red}{red} bars represent results trained with \texttt{DeepSeek-R1}/\texttt{gpt-oss-120b} generated reasoning trajectories, respectively.}
    \label{fig:random_del_reason_step_appendix}
\end{figure}

To explore more about the random reasoning step deletion experiments, we adopt different deletion ratio (i.e., delete 10\%, 20\%, and 30\% reasoning steps from each long CoT trajectories).
The experiment results with Qwen3-8B are shown in Figure~\ref{fig:random_del_reason_step_appendix}: our observation that performance degradation of \texttt{gpt-oss-120b}-distilled models is much more drastic than \texttt{DeepSeek-R1-0528}-distilled models consistently holds for all three deletion ratios.
We observe that when randomly deleting 20\% steps from each \texttt{DeepSeek-R1} reasoning trajectory, the re-trained Qwen3-8B's performance on BeyondAIME (\texttt{avg@10}) even increases by relatively 5.0\% (normalized by the original performance). 
These results together validate our observation on the redundancy of \texttt{DeepSeek-R1} reasoning trajectories again.

\subsection{Additional Results of Improving SFT through Filtering out the Most Frequently Branching Trajectories}
\label{appendix:improve_sft_extend_to_32k}
We provide more experiment results on re-training Qwen2.5-7B after filtering out part of the frequently branching SFT training data, where we set the context limit = 32K at inference time. The results are shown in Table~\ref{tab:improve_sft_extend_to_32k}, which support the key conclusion presented in the main text.
\begin{table}[ht]
\centering
\caption{More experiment results on filtering out part of the frequently branching SFT training data with Qwen2.5-7B, where \textbf{we set the context limit = 32K at inference time.}}
\label{tab:improve_sft_extend_to_32k}
\resizebox{\textwidth}{!}{%
\begin{tabular}{@{}l|c|ccccc|c@{}}
\toprule
\textbf{Base Model} & \textbf{Data Source} & \textbf{MATH500} & \textbf{AIME24} & \textbf{AIME25} & \textbf{BeyondAIME} & \textbf{HMMT25} & \textbf{Avg} \\
\midrule
\multirow{9}{*}{\shortstack{Qwen2.5-7B \\ {\footnotesize (context len: 32K)}}} 
 & full \texttt{DeepSeek-R1} set (original SFT) & $95.0\%$ & $52.1\%$ & $44.8\%$ & $16.7\%$ & $38.7\%$ & $49.5\%$ \\
  \cmidrule{2-8}
 & removing top $10\%$& $95.0\%$ & \cellcolor{green!25}$54.3\%$ & \cellcolor{green!25}$46.7\%$ & \cellcolor{green!25}$19.3\%$ & \cellcolor{green!25}$40.2\%$ & $\mathbf{51.1\%}$ {\footnotesize \textcolor{green!60!black}{(\textbf{+1.6\%})}}\\
 & removing top $20\%$& \cellcolor{green!25}$95.8\%$ & \cellcolor{green!25}$52.2\%$ & \cellcolor{red!25}$44.4\%$ & \cellcolor{green!25}$16.2\%$ & \cellcolor{green!25}$40.0\%$ & $\mathbf{49.7\%}$ {\footnotesize \textcolor{green!60!black}{(\textbf{+0.2\%})}}\\
   \cmidrule{2-8}
 & \texttt{bottom $\mathbf{90\%}$ (experiment group)} & \cellcolor{green!25}$95.4\%$ & \cellcolor{green!25}$53.7\%$ & \cellcolor{green!25}$46.5\%$ & \cellcolor{green!25}$19.3\%$ & \cellcolor{green!25}$41.8\%$ & $\mathbf{51.3\%}$ {\footnotesize \textcolor{green!60!black}{(\textbf{+1.8\%})}} \\
  & \texttt{top $\mathbf{90\%}$ (control group)}& $95.0\%$ & \cellcolor{red!25}$51.9\%$ & \cellcolor{green!25}$44.9\%$ & \cellcolor{green!25}$17.7\%$ & \cellcolor{red!25}$37.1\%$ & $49.3\%$ {\footnotesize \textcolor{red!60!black}{(\textbf{-0.2\%})}} \\
\cmidrule{2-8}
 & \texttt{bottom $\mathbf{80\%}$ (experiment group)} & \cellcolor{red!25}$94.6\%$ & \cellcolor{green!25}$55.5\%$ & \cellcolor{green!25}$47.2\%$ & \cellcolor{green!25}$18.5\%$ & \cellcolor{green!25}$40.1\%$ & $\mathbf{51.2\%}$ {\footnotesize \textcolor{green!60!black}{(\textbf{+1.7\%})}} \\
  & \texttt{top $\mathbf{80\%}$ (control group)} & \cellcolor{red!25}$94.4\%$ & \cellcolor{red!25}$52.0\%$ & \cellcolor{red!25}$44.5\%$ & \cellcolor{red!25}$15.0\%$ & \cellcolor{green!25}$39.3\%$ & $49.0\%$ {\footnotesize \textcolor{red!60!black}{(\textbf{-0.5\%})}} \\
 \cmidrule{2-8}
 & \texttt{bottom $\mathbf{50\%}$ (experiment group)}  & \cellcolor{green!25}$95.4\%$ & \cellcolor{red!25}$51.9\%$ & \cellcolor{green!25}$46.0\%$ & \cellcolor{green!25}$19.1\%$ & \cellcolor{green!25}$41.0\%$ & $\mathbf{50.7\%}$ {\footnotesize \textcolor{green!60!black}{(\textbf{+1.2\%})}} \\
  & \texttt{top $\mathbf{50\%}$ (control group)} & \cellcolor{red!25}$94.6\%$ & \cellcolor{red!25}$50.6\%$ & \cellcolor{red!25}$43.2\%$ & \cellcolor{red!25}$15.0\%$ & \cellcolor{red!25}$38.3\%$ & $48.3\%$ {\footnotesize \textcolor{red!60!black}{(\textbf{-1.2\%})}} \\
\bottomrule
\end{tabular}
}
\end{table}

\section{Case Study: Highly Exploratory Reasoning Trajectories Generated by \texttt{DeepSeek-R1-0528} and Its Distilled Student Model}
\label{appendix:long_cot_showcase}
This section contains two specific reasoning trajectory examples (snippets): one is a highly exploratory (repetitively branching) snippet of a reasoning trajectory generated by \texttt{DeepSeek-R1-0528} (in the SFT training set, Figure~\ref{figure:showcase_training_data}); another is a highly exploratory (repetitively branching) snippet of a reasoning trajectory generated by the Qwen3-8B that was trained on \texttt{DeepSeek-R1-0528} trajectories (a generated solution for a evaluation problem, Figure~\ref{figure:showcase_testing_data}). 
We can observe that the Qwen3-8B base model after SFT inherits the frequently branching reasoning pattern in the \texttt{DeepSeek-R1} reasoning trajectories, which potentially hinder the trained model from reaching the correct solutions.

\begin{figure*}[ht]
\centering
\begin{tcolorbox}[width=0.97\textwidth, colback=white,colframe=black,title=A reasoning trajectory generated by \texttt{DeepSeek-R1-0528}]
\scriptsize
\textbf{USER:}

Find the smallest positive integer n such that $3^n \equiv 1 \pmod{1000000007}$.

\vspace{1em}

\textbf{ASSISTANT:}

\vspace{1em}

\textit{\textbf{...... (omitting previous reasoning steps)}}

\vspace{1em}

\textcolor{red}{Perhaps} 1000000006 = 2 * 500000003, and 500000003 = 7 * 71428571.857, but not, \textcolor{red}{or perhaps} 71428571 is the key. \textbf{('Propose')}

\vspace{1em}

71428571 * 7 = 499,999,997, as before. \textbf{('Deduce')}

\vspace{1em}

Then 71428571, let's see if it's divisible by 7: 7*10204081.571, not. \textbf{('Deduce')}

\vspace{1em}

By 11: 7-1+4-2+8-5+7-1=7-1=6, +4=10, -2=8, +8=16, -5=11, +7=18, -1=17 not div by 11. \textbf{('Deduce')}

\vspace{1em}

By 13: 13*5494505.4615, 13*5494505 = 13*5000000=65,000,000, 13*494505=6,428,565, total 71,428,565, as before, 71,428,571-71,428,565=6, not div. \textbf{('Deduce')}

\vspace{1em}

\textcolor{red}{Perhaps} 71428571 = 2381 * 30000, but 2381*30000=71,430,000 $>$ 71,428,571. \textbf{('Propose')}

\vspace{1em}

381*29999 = 2381*30000-2381=71,430,000-2,381=71,427,619, then 71,428,571-71,427,619=952, not div. \textbf{('Deduce')}

\vspace{1em}

\textcolor{red}{Perhaps} it is 6449*11079 \textcolor{red}{or something}. \textbf{('Propose')}

\vspace{1em}

I found online that 1000000006 = 2 * 500000003, and 500000003 = 7 * 71428571, but 7*71428571=499999997, not 500000003. \textbf{('Deduce')}

\vspace{1em}

500000003 / 7 = 71428571.857..., not integer. \textbf{('Deduce')}

\vspace{1em}

\textcolor{red}{Perhaps} 500000003 = 641 * 780000.472, not. \textbf{('Propose')}

\vspace{1em}

Let's try 641. 641*780000=499,800,000, 641*342=219,222, total 500,019,222, too big. \textbf{('Deduce')}

\vspace{1em}

641*779000=499,339,000, 641*342=219,222, total 499,558,222, then 500,000,003-499,558,222=441,781, not div. \textbf{('Deduce')}

\vspace{1em}

This is taking too long. \textcolor{red}{Perhaps} for the sake of the problem, the order is p-1, but that is not smallest. \textbf{('Propose')}

\vspace{1em}

\textcolor{red}{Another idea:} perhaps 3 is a primitive root modulo 1000000007, so the order is p-1 = 1000000006. \textbf{('Propose')}

\vspace{1em}

But that would make n large, and probably not, since the problem asks for smallest. \textbf{('Verify')}

\vspace{1em}

\textcolor{red}{Perhaps} I can compute $3^{(p-1)/2}$ mod p to see if it is -1 or 1. \textbf{('Propose')}

\vspace{1em}

By Euler's criterion, since p is prime, $3^{(p-1)/2}$ $\equiv$ (3/p) mod p, where (3/p) is the Legendre symbol. \textbf{('Deduce')}

\vspace{1em}

So if (3/p) = 1, then order divides (p-1)/2, else not. \textbf{('Deduce')}

\vspace{1em}

\textit{\textbf{...... (omitting latter reasoning steps)}}

\vspace{1em}

Thus the answer is $\backslash$boxed\{500000003\}.
\end{tcolorbox}
\caption{Case study: a highly exploratory snippet of a reasoning trajectory generated by \texttt{DeepSeek-R1-0528}.}
\label{figure:showcase_training_data}
\end{figure*}

\begin{figure*}[ht]
\centering
\begin{tcolorbox}[width=0.97\textwidth, colback=white,colframe=black,title=A reasoning trajectory generated by \texttt{DeepSeek-R1-0528}-distilled Qwen3-8B]
\scriptsize

\vspace{1em}

\textit{\textbf{...... (omitting previous reasoning steps)}}

\vspace{1em}

\textcolor{red}{Perhaps} for the specific quadrant, I can solve numerically, but the answer is given as $(a-\sqrt{b})/c$, so probably rational. \textbf{('Propose')}

\vspace{1em}

\textcolor{red}{Perhaps} the unique point is when the rotation maps to itself or something, but unlikely. \textbf{('Propose')}

\vspace{1em}

\textcolor{red}{Another thought: perhaps} the point is such that after rotation, it is the same point, but only origin. \textbf{('Propose')}

\vspace{1em}

\textcolor{red}{Or perhaps} for the parabola, the intersection occurs when the point is at 60 degrees from the axis. \textbf{('Propose')}

\vspace{1em}

Let's think geometrically. \textbf{('Deduce')}

\vspace{1em}

The original parabola $y = x^2 -4$, vertex at (0,-4), opens up. \textbf{('Deduce')}

\vspace{1em}

Rotated by 60 degrees counterclockwise around origin. \textbf{('Deduce')}

\vspace{1em}

The rotated parabola will intersect the original at some points. \textbf{('Deduce')}

\vspace{1em}

Since it's symmetric about y-axis, and rotation by 60 degrees, which is not symmetric, so intersections may not be symmetric. \textbf{('Deduce')}

\vspace{1em}

But the problem says unique in fourth quadrant, so probably two in first and two in fourth, but unique in fourth, so perhaps only one in fourth, but due to symmetry, should be two, unless on axis. \textbf{('Verify')}

\vspace{1em}

Fourth quadrant is $x>0$, $y<0$, so should be two points, but the problem says unique, so \textcolor{red}{perhaps} only one, \textcolor{red}{or perhaps} on the axis, but $y<0$, so not on x-axis. \textbf{('Propose')}

\vspace{1em}

\textcolor{red}{Perhaps} at $x=0$, but $y=-4$, not on rotated. \textbf{('Propose')}

\vspace{1em}

Let's calculate the y-coordinate from the expression. \textbf{('Deduce')}

\vspace{1em}

\textcolor{red}{Perhaps} the y-coordinate is given, and it's $(a-\sqrt{b})/c$, so \textcolor{red}{perhaps} it's a root of a quadratic. \textbf{('Propose')}

\vspace{1em}

\textit{\textbf{...... (omitting latter reasoning steps)}}

\vspace{1em}

\end{tcolorbox}
\caption{Case study: a highly exploratory snippet of a reasoning trajectory generated by \texttt{DeepSeek-R1-0528}-distilled Qwen3-8B.}
\label{figure:showcase_testing_data}
\end{figure*}

\end{document}